\documentclass[acmlarge, nonacm, screen]{acmart}
\usepackage{algorithm}
\usepackage{algorithmic}
\usepackage{xcolor}

\usepackage{subfig}
\usepackage{booktabs}

\usepackage{siunitx}
\usepackage{multirow}

\usepackage{newfloat}
\usepackage{listings}

\setcopyright{none}
\renewcommand\footnotetextcopyrightpermission[1]{} % removes footnote with conference information in first column
\authorsaddresses{%
  \vspace{1ex}%
  \noindent
    Corresponding Author: Rebecca Dorn, rdorn@usc.edu
}

\AtBeginDocument{%
  }

\setcopyright{acmlicensed}
\copyrightyear{2025}
\acmYear{2025}
\acmDOI{XXXXXXX.XXXXXXX}
\acmConference[Conference acronym 'XX]{Make sure to enter the correct
  conference title from your rights confirmation email}{June 03--05,
  2018}{Woodstock, NY}
\acmISBN{978-1-4503-XXXX-X/2018/06}

\begin{document}

%%
%% The "title" command has an optional parameter,
%% allowing the author to define a "short title" to be used in page headers.
\title{Reinforcing Stereotypes of Anger: Emotion AI on African American Vernacular English}

%%
%% The "author" command and its associated commands are used to define
%% the authors and their affiliations.
%% Of note is the shared affiliation of the first two authors, and the
%% "authornote" and "authornotemark" commands
%% used to denote shared contribution to the research.

\author{Rebecca Dorn}
\email{rdorn@usc.edu}
\affiliation{%
  \institution{University of Southern California, Information Science Institute}
  \city{Marina del Rey}
  \state{CA}
  \country{USA}
}

\author{Christina Chance}
\email{cchance@cs.ucla.edu}
\affiliation{%
  \institution{University of California, Los Angeles}
  \city{Los Angeles}
  \state{CA}
  \country{USA}
}

\author{Casandra Rusti}
\email{rusti@usc.edu}
\affiliation{%
  \institution{University of Southern California, Information Science Institute}
  \city{Marina del Rey}
  \state{CA}
  \country{USA}
}

\author{Charles Bickham Jr.}
\email{cbickham@usc.edu}
\affiliation{%
  \institution{University of Southern California, Information Science Institute}
  \city{Marina del Rey}
  \state{CA}
  \country{USA}
}

\author{Kai-Wei Chang}
\email{kwchang@cs.ucla.edu}
\affiliation{%
  \institution{University of California, Los Angeles}
  \city{Los Angeles}
  \state{CA}
  \country{USA}
}

\author{Fred Morstatter}
\email{fredmors@isi.edu}
\affiliation{%
  \institution{University of Southern California, Information Science Institute}
  \city{Marina del Rey}
  \state{CA}
  \country{USA}
}

\author{Kristina Lerman}
\email{lerman@isi.edu}
\affiliation{%
  \institution{University of Southern California, Information Science Institute}
  \city{Marina del Rey}
  \state{CA}
  \country{USA}
}

%%
%% By default, the full list of authors will be used in the page
%% headers. Often, this list is too long, and will overlap
%% other information printed in the page headers. This command allows
%% the author to define a more concise list
%% of authors' names for this purpose.
\renewcommand{\shortauthors}{Dorn et al.}

%%
%% The abstract is a short summary of the work to be presented in the
%% article.
\begin{abstract}
Automated emotion detection is widely used in applications ranging from product reviews and well-being monitoring, to high-stakes domains like mental health and hiring. 
However, models often rely on annotations that reflect dominant cultural norms, limiting
 model ability to recognize emotional expression in dialects often excluded from training data distributions, such as African American Vernacular English (AAVE).
%This reliance limits model ability to recognize emotional expression in linguistically diverse communities, such as speakers of African American Vernacular English (AAVE).
This study examines how emotion recognition models perform on AAVE in comparison to General American English (GAE).
%\katie{GAE is probably the most common term, but in recent years, there's bene a push to rename the dialect as Mainstream American English (GAE) or General American English (GAE) in recognition that the word "standard" comes with a prescriptivist notion of it being the "correct" or "proper" dialect, which is sociologically governed, rather than simply discussing this dialect of English as linguistically equivalent to any other dialect. As GAE is still more common and recognizable, you may want to keep that term, but it may be worth acknowledging the other terms, too.}. 
We analyze 2.7 million tweets geo-tagged within the Los Angeles area. Each text is scored for strength of AAVE using computational approximations of dialect features. 
Annotations of emotion presence and intensity are collected on a dataset of 875 tweets with both high and low AAVE densities.
%To assess model accuracy, we garner annotations from 11 individuals for emotion presence and intensity on a subset of 875 tweets, featuring both high and low densities of AAVE features. 
To assess model accuracy on a task as subjective as emotion perception, we calculate community-informed ``silver" labels where AAVE-dense tweets are labeled solely by African American, AAVE-fluent (ingroup) annotators.
% giving more weight to community members when generating semi-ground truth labels for high-AAVE texts.
% Overall, AAVE-authored tweets are disproportionately labeled as angry, while GAE tweets are portrayed as more joyful and less angry. 
On our labeled sample, GPT and BERT-based models exhibit false positive prediction rates of anger on AAVE more than double than on GAE.
SpanEmo, one of the most popular text-based emotion AI models, increases its false positive rate of anger from 25\% on GAE to 60\% on AAVE.
Additionally, a series of linear regressions reveals that models and non-ingroup annotations are significantly more correlated with profanity-based AAVE features than ingroup annotations.
% Our results reveal significant performance disparities: emotion detection models are more accurate at detecting anger in high-AAVE tweets but consistently under-predict joy. GAE tweets, by contrast, produce more false positives for joy and false negatives for anger. 
Linking tweets to Census tract demographics, we observe that neighborhoods with higher proportions of African American residents are associated with higher predictions of anger (Pearson’s correlation $r = 0.27$, p < 0.01) and lower joy ($r = -0.10$, p < 0.01). 
These results find an emergent safety issue of emotion AI reinforcing racial stereotypes through biased emotion classification. We emphasize the need for culturally and dialect-informed affective computing systems.\\
%These results highlight the risk of reinforcing racial stereotypes through biased emotion classification and emphasize the need for culturally and dialect-informed affective computing systems.\\
\textcolor{red}{\textit{Warning: This paper contains discussions of slur usage and discriminatory experiences.}}
\end{abstract}

%%
%% The code below is generated by the tool at http://dl.acm.org/ccs.cfm.
%% Please copy and paste the code instead of the example below.
%%
\begin{CCSXML}
<ccs2012>
   <concept>
       <concept_id>10003120</concept_id>
       <concept_desc>Human-centered computing</concept_desc>
       <concept_significance>500</concept_significance>
       </concept>
   <concept>
       <concept_id>10003456.10010927.10003611</concept_id>
       <concept_desc>Social and professional topics~Race and ethnicity</concept_desc>
       <concept_significance>500</concept_significance>
       </concept>
   <concept>
       <concept_id>10010147.10010257</concept_id>
       <concept_desc>Computing methodologies~Machine learning</concept_desc>
       <concept_significance>500</concept_significance>
       </concept>
   <concept>
       <concept_id>10002951.10003260.10003282.10003292</concept_id>
       <concept_desc>Information systems~Social networks</concept_desc>
       <concept_significance>500</concept_significance>
       </concept>
 </ccs2012>
\end{CCSXML}

\ccsdesc[500]{Human-centered computing}
\ccsdesc[500]{Social and professional topics~Race and ethnicity}
\ccsdesc[500]{Computing methodologies~Machine learning}
\ccsdesc[500]{Information systems~Social networks}

\keywords{Affective Computing, Social Networks, Algorithm Audit}

\received{20 February 2007}
\received[revised]{12 March 2009}
\received[accepted]{5 June 2009}

%%
%% This command processes the author and affiliation and title
%% information and builds the first part of the formatted document.
\maketitle

\section{Introduction}

Emotion AI use is growing, ranging from mental health and therapy chat-bots \cite{andrade2024therapy} to public opinion and organizing involvement in the \textit{Black Lives Matter} movement ~\cite{patnaude2021public,vanHaperen2022swarm}. 
These emotion recognition systems are applications of Natural Language Processing (NLP), where models are typically trained on data reflecting dominant language norms, namely General American English (GAE) \footnote{In this paper, General American English (GAE) refers to the American dialect missing noticeable ethnic or region markers that is frequently represented in media and academia \cite{van1986general, gramley2020survey}.}.
Consequently, these systems may struggle to accurately interpret emotional expressions in dialects spoken by historically marginalized communities, such as African American Vernacular English (AAVE). AAVE has been formalized as a systematic, rule-governed variety of English with distinct lexical and syntactic features, yet it is often misclassified by NLP systems. For instance, Zoom’s closed captioning service performs significantly worse on AAVE~\cite{chance2022zoom}, and widely used toxicity detectors disproportionately flag AAVE speech as offensive~\cite{halevy2021mitigating}. As affective computing systems are built on NLP models, there is growing concern that they too will misinterpret the tone, affect, and intent of AAVE speakers, potentially amplifying harmful stereotypes and undermining the reliability of emotion AI. 

This study investigates how emotion recognition models respond to AAVE, with a specific focus on individual sociolinguistic feature influence on automated affect. 
In this work, we use \textit{AAVE} to refer to the frequently shared lexical, orthographic 
%\katie{how would you captured phonological features in text? You might consider swapping this for lexical if it's word-driven or orthographic if there's a particular AAVE spelling associated with an GAE word}
and grammatical features of English spoken in African American communities, such as habitual ``be'' ("They \textit{be} the real troublemakers" \cite{alim2006roc, spears2019rickford}) and negative concord (e.g., "He \textit{ain't} got \textit{no} car" \cite{Martin_Wolfram_2021}) ~\cite{aave_oxford, Martin_Wolfram_2021, Smitherman_2021, Green_2021}. 
%the Los Angeles variant of AAVE, with a specific focus on individual sociolinguistic feature influence on automated affect detection. 
This work analyzes 2.7 million geo-tagged tweets from the greater Los Angeles area. Texts are scored for intensity of AAVE linguistic features using computational proxies for both national and Los Angeles-specific dialect features. %\katie{Cool consideration regarding LA regional dialect, but you might give an example of an LA-specific regional dialectal feature that differs from GAE (and that would be present in text format as opposed to phonological pronunciation).} 
To ground the evaluation, we collect labels  on a sample 875 tweets that is balanced between high and low amounts of AAVE from 11 annotators. 
We determine community-informed ``silver" labels where tweets with higher amounts of AAVE are judged only by African American, AAVE-fluent (hereafter \textbf{ingroup}) annotators. Tweets with low or no AAVE features are labeled by all annotators.
These tweets provide a rich dataset for examining how AAVE features influence emotion detection across lexicon-based and transformer-based emotion models.
%To ground the evaluation, we collect emotion annotations from 11 annotators, including African American Human labels are collected from 11 ann
%both AAVE-fluent community members and non-community annotators. This allows us testablish ``semi-ground truth" emotion labels, with greater weight given to community members for labeling tweets with high densities of AAVE features.
%\kl{Why "semi-ground truth"? I thought we collect gold labels from community members. Did we introduce "AAVE feature" density yet? Is there a way to keep the discussion more general by avoiding "feature density" language?} 
Our research questions are as follows:
\begin{description}
\item[RQ1] How does group membership influence emotion annotations of AAVE and GAE texts?
\item[RQ2] How do emotion AI models perform on AAVE text compared to GAE text? Which annotation groups do model predictions align with?
\item[RQ3] Which linguistic features of AAVE correlate with model performance?
\item[RQ4] Do linguistic biases lead AI models to infer systematically different emotions in predominantly Black and White neighborhoods?
\end{description}

We find that annotations are mediated by group membership. Specifically, ingroup members identify more emotion in all tweets except for disgust on high AAVE tweets, where outgroup members label excess.
On this labeled sample, models falsely predict anger and disgust in AAVE at rates more than double those in GAE. For the popular BERT-based model \textit{SpanEmo}, the false positive rate on anger rockets from 25\% on GAE to 60\% on AAVE. 
Looking at the influence of individual features of AAVE on labels, compared to labels from ingroup annotators, both model and outgroup annotations are both significantly more influenced by profanity-based features of AAVE.
These results underscore the need for caution in deploying emotion recognition systems, highlighting the importance of considering the cultural and linguistic patterns of the target population.

\begin{figure*}[!htp]
  \centering
  \subfloat[Percent African American.]{\includegraphics[width=0.33\textwidth]{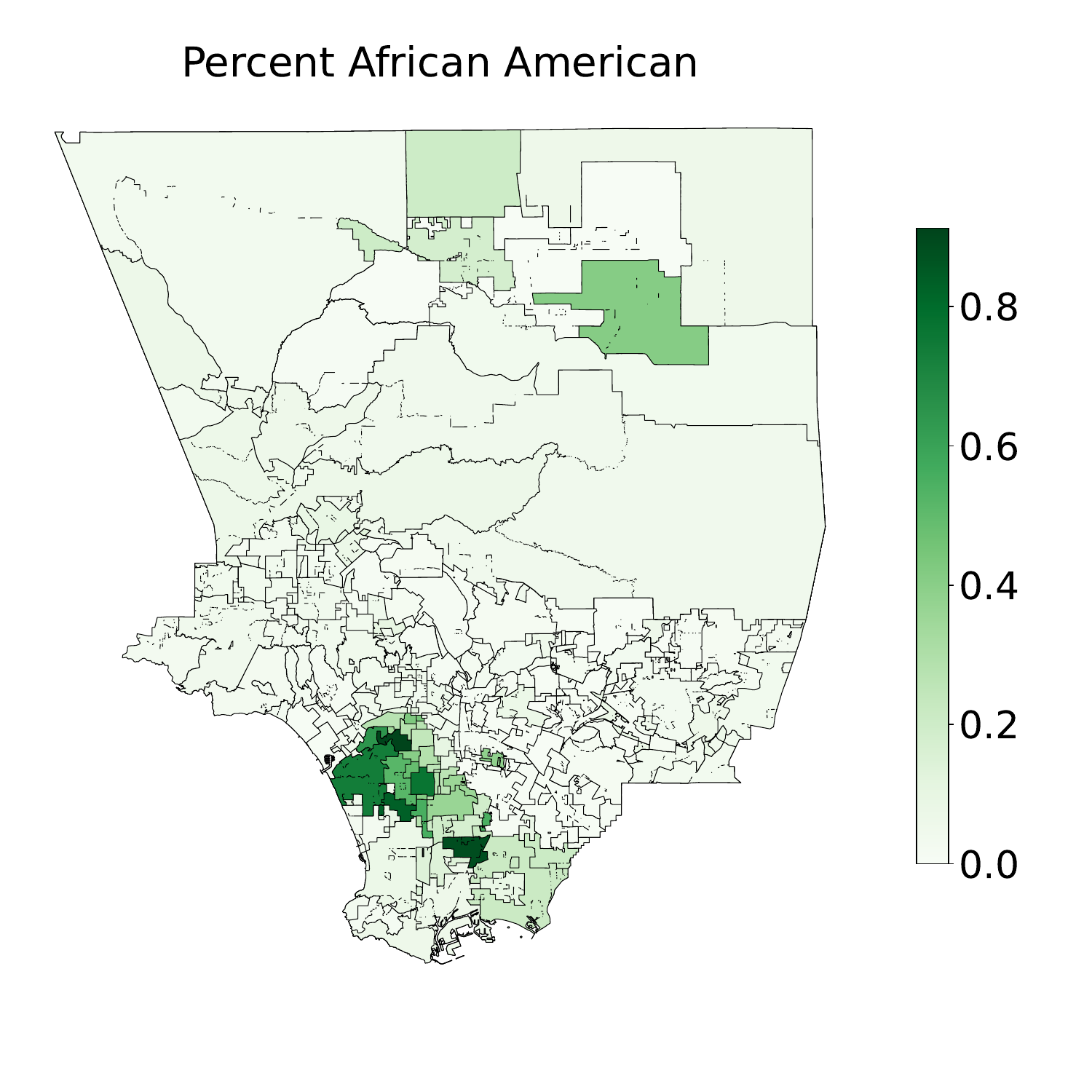}\label{fig:f1}}
  \hfill
  \subfloat[AAVE Dialect Density.]{\includegraphics[width=0.33\textwidth]{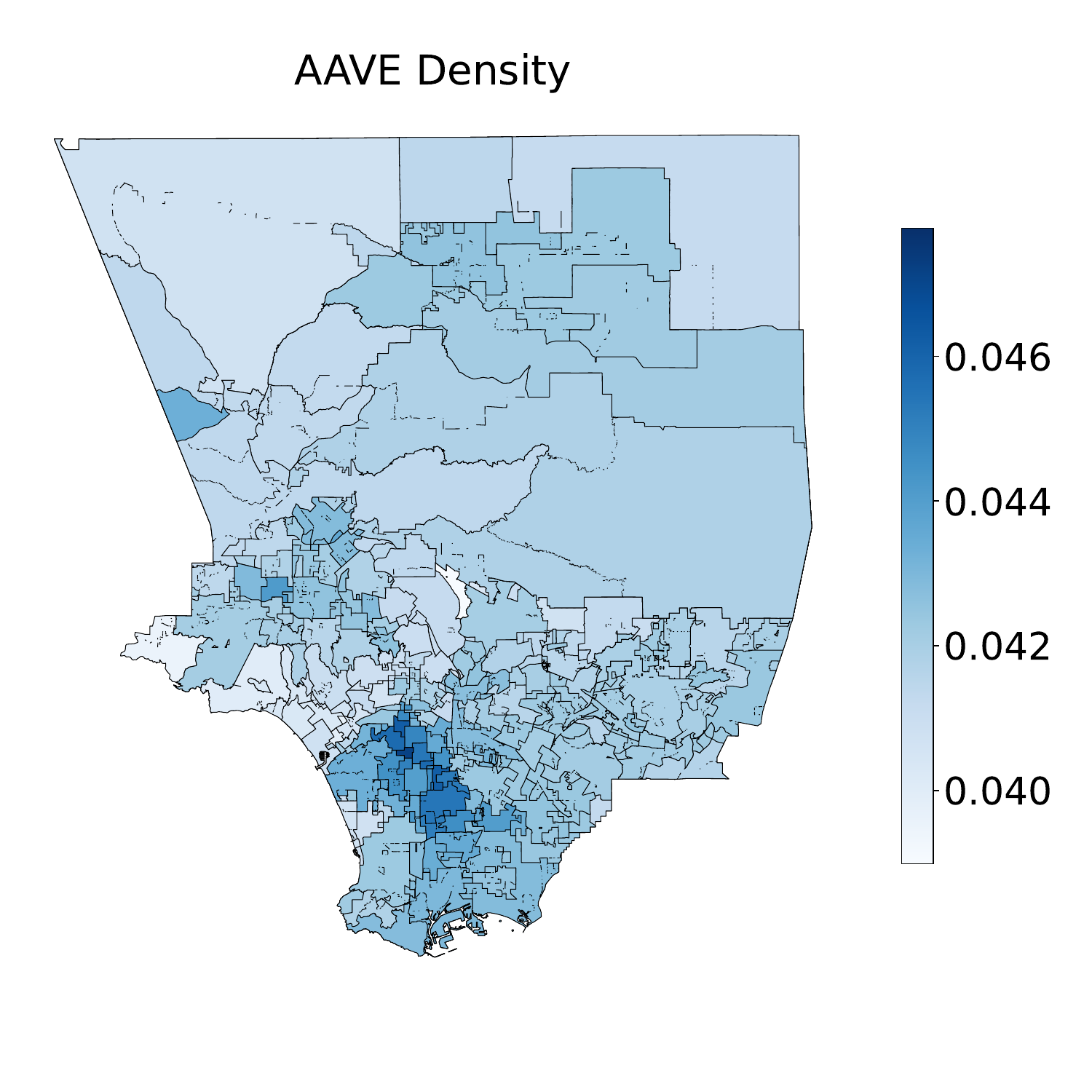}\label{fig:f2}}
  \hfill
  \subfloat[SpanEmo P(Anger)]{\includegraphics[width=0.33\textwidth]{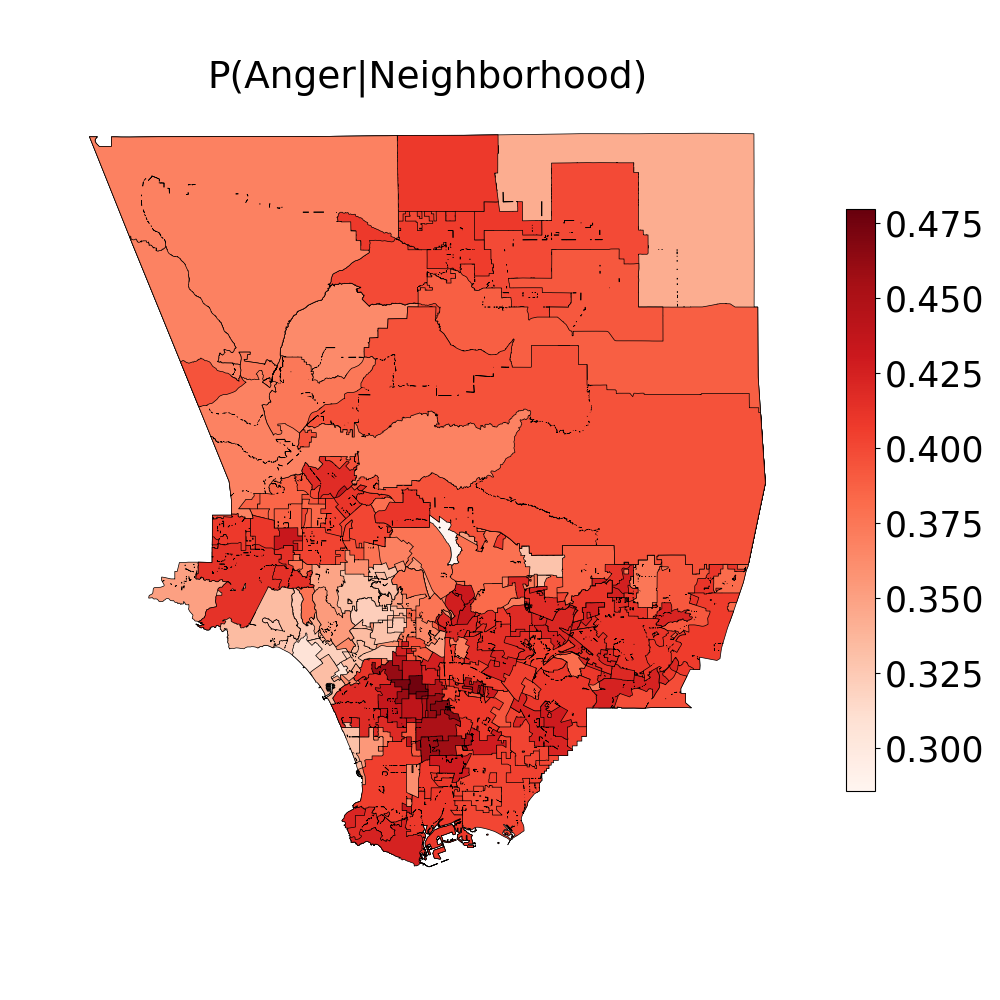}\label{fig:f3}}
\caption{Relationship between neighborhood demographics, dialect density, and predictions of anger from SpanEmo, a popular BERT-based model. Neighborhood boundaries are calculated are determined using groupings of Census tracts outlined by \textit{Mapping LA} \cite{latimesMappingLA}. Demographics are approximated from Census.}
\end{figure*}

\section{Related Work}

\subsection{Computing Emotion}
Measuring emotion requires condensing abstract conceptions of emotions into discrete constructs. 
Computational spaces approach this challenge through either categories of emotions or probability values for specific affect aspects, such as arousal or valence ~\cite{arco2024,Zad_Heidari_Jones_Uzuner_2021,Calvo2012emotions}.
We focus on the four main categorical emotion conceptualizations that are typically used in contemporary affective computing: Ekman, Plutchik, Parrot, and GoEmotions ~\cite{arco2024,Zad_Heidari_Jones_Uzuner_2021,Calvo2012emotions}.
Ekman's foundational work used facial expressions to establish six basic emotions (joy, sadness, surprise, disgust, anger and fear) ~\cite{ekman1978facial, Ekman1992-EKMAAF}. 
Shortly after, Plutchik used a psycho-evolutionary lens to expanded Ekman's list to include anticipation and trust~\cite{plutchik1980emotion}. 
Decades later, a multi-layer hierarchical schema with six primary categories was proposed by Parrott ~\cite{parrott2001emotions, murthy2021review}.
Recently, an analysis into emotional states elicited from short videos introduced GoEmotions, a framework of 27 emotions widely adopted in emotion AI \cite{demszky2020goemotions, cowen2017self}. 
While these discrete frameworks provide essential structure for emotion detection tasks, their limitations warrant consideration. The widespread application of category-based models across diverse datasets often disregards cultural nuances in emotional expression and recognition~\cite{santos2018language}. 
This uncritical adoption of universal emotion theories can yield biased or inaccurate results, underscoring the need for more culturally sensitive approaches in emotion AI~\cite{santos2018language}.

\subsection{African American Affect Expression \& Perception}
Emotion expression and perception are intricately mediated by culture, including race \cite{Wilson_Gentzler_2021, weiss2022racial}.
Some African American communities have distinct emotion expression norms for articulating affect compared to White American communities.
 For example, ``You're a fool'' may be interpreted by White audiences as an insult, but within African American communities the phrase often indicates banter or play     \cite{doi:10.1177/1527476413480247}.
 These norms are context-dependent, influenced by code-switching \cite{powell2023switch}, community support, discriminatory experiences, and the intended audience \cite{Wingfield_2010, Brown_Lozada_Serpell_Dzokoto_Dunsmore_2025}. 
Regarding emotion perception, sociodemographic group membership significantly impacts the accuracy of emotion interpretation. White individuals often misinterpret African American emotion expression, mistakenly reading positive affect as negative or interpreting ambiguous facial expressions as anger \cite{Wickline_Bailey_Nowicki_2009, Rhue_2018}.
This ``angry black person" stereotype perpetuates racial discrimination, resulting in increased health risks and disproportionate law enforcement in schools with high concentrations of African American students \cite{lozada2022black, walley2009debunking}. 
This work aims to mitigate the proliferation of these misjudgments by having texts written in the features of African American Vernacular English evaluated solely by African American annotators. 

\subsection{African American Vernacular English (AAVE)}
In this work, we use AAVE to refer to the frequently shared lexical, orthographic
and grammatical features of English spoken in African American communities. 
While many of these features are common across North American English dialects, AAVE exhibits meaningful regional variation, shaped by historical migration patterns. 
Los Angeles has served as the focal point to AAVE's academic formulation ~\cite{Rickford}, and the city's entertainment industry plays a strong role in representations in film and television. 
Further, Los Angeles has a unique variant due to Southern California's relationship with Mexican communities influencing local speech patterns ~\cite{Jones_2015a, great_migration_la, Rickford}. 
To allow for integration with Census demographics and enable regionally-tailored computations of AAVE, we add some features tied to the Los Angeles variant of AAVE such as the \textit{continuative steady} (e.g., ``Marcus \textit{steady} trying to get with Erica"" \cite{knapp2015african}) which was first studied in Black Californian communities \cite{baugh_linguistic_1979}.% \katie{you might consider using some examples} 
%The Great Migration and Second Great Migration brought African Americans to Southern California, where contact with White and Mexican communities influenced local speech patterns~\cite{Jones_2015a, great_migration_la, aave_california}.  %\katie{Did have a comment here that the migration bit felt a little disconnected, so taking it out is a good thing, I think.}

\subsection{AAVE Treatment in NLP \& Emotion Recognition}

Research has consistently shown that NLP and machine learning tools exhibit biases against AAVE across tasks from automated speech recognition to natural language understanding \cite{deas-etal-2023-evaluation, gupta-etal-2024-aavenue, martin20_interspeech, groenwold-etal-2020-investigating, blodgett2017racialdisparitynaturallanguage}. However, fewer works examine how these biases impact automated emotion recognition.
In contrast, social sciences literature has  extensively  documented the misrepresentation  of emotions expressed by Black individuals \cite{doi:10.1111/j.0956-7976.2004.00680.x, halberstadt2022racialized, friesen2019perceiving}. These systemic misunderstandings risk being embedded into development and deployment of emotion recognition models, perpetuating harms.

\subsection{Dialect Density}

\begin{figure*}[h]
\centering
\includegraphics[width=0.9\textwidth]{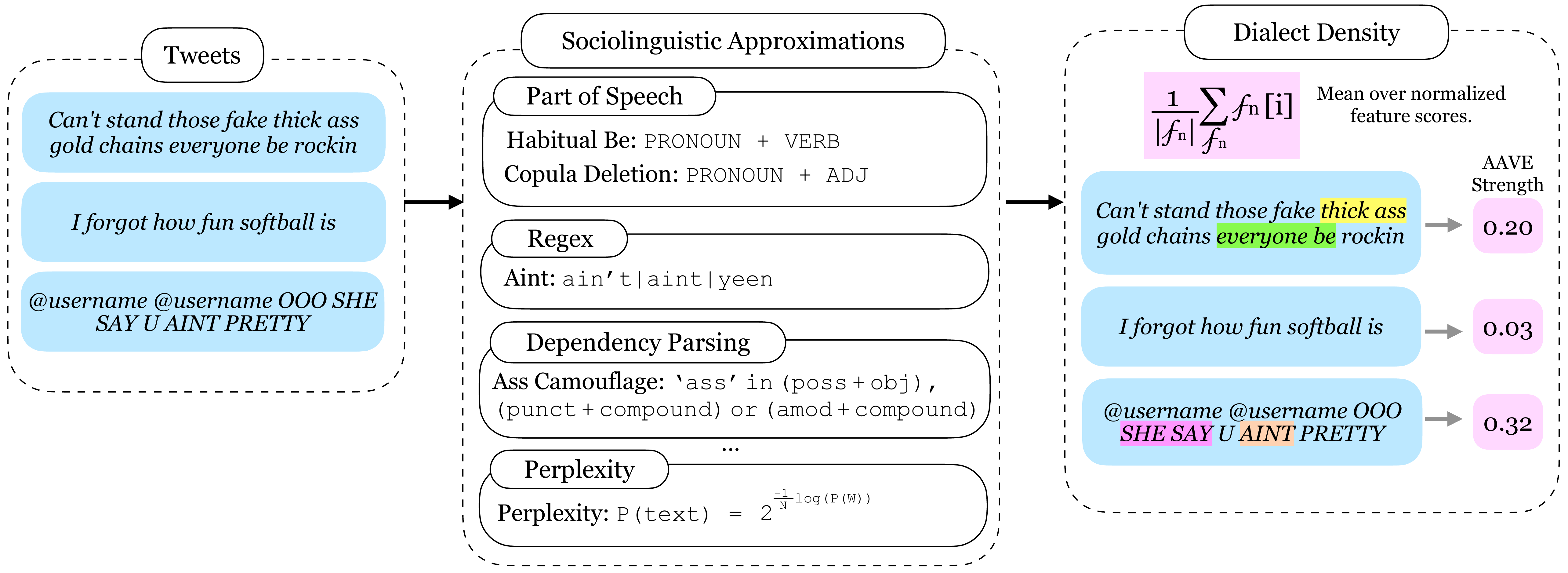}
\caption{Method to approximate AAVE Dialect (DDM) in a text. Sociolinguistic features are individually approximated using NLP tools like regex expressions, dependency parsing, named entity recognition and perplexity. Taking the average over normalized features yields scalar score \texttt{DDM}.}
\label{fig:ddmmap}
\end{figure*}

As opposed to representing dialect presence using a binary variable, \textit{dialect density} approximates the strength a sociolinguistic dialect features in a text. 
Previous work exploring AAVE dialect density, tweet locations, and surrounding demographic information revealed that predominantly African American areas often display established phonological and syntactical features of AAVE \cite{Blodgett_Green_O’Connor_2016}.  Building on this, \citet{Johnson2022} introduced a method to automatically assess AAVE dialect density from audio, finding a strong correlation between their dialect density metric (DDM) and speaker demographics. We adapt select features of this method to calculate the dialect density of tweets, as outlined in  Methods \ref{sec:ddm}. 
We use DDM as a proxy for AAVE-speaker classification, acknowledging that not all  African-Americans speak in AAVE and that AAVE features are adopted by other sociodemographic groups (e.g., queer communities \cite{crowley2025and}).

\section{Methods}
%DONE -- \cas{Data, Emotion Operationalization, DD, Annotation, Model Choice}

\subsection{Dataset}
\paragraph{Tweet Collection and Filtering.} We use a dataset filtered from 5.8 million geo-tagged tweets posted within Los Angeles County, spanning September 2010 to November 2014. 
Though the dataset is older than typical machine learning datasets, this time-frame is a frequent study site of \textit{black twitter} (e.g. \cite{Jones_2015a}), as it occurred before Twitter's change in leadership.
%\katie{Why is this time-frame a frequent study site?}
We drop short tweets ($\leq$ 5 tokens) as they inherently do not have enough text to estimate emotion and AAVE density. We exclude tweets with links to lessen the need for external context. 
For dataset cleaning, usernames are anonymized such that username tags are swapped with \texttt{@username}, emojis are converted to their text descriptors, and all URLS are removed. 
This results in 2.7M tweets for analysis. 

\paragraph{Neighborhood and Demographic Mapping.}
The dataset is enriched with demographic and neighborhood-level metadata.
Tweet coordinates are mapped to its 2010 Census tract and joined with population statistics from the 2014 LA County dataset \cite{lacounty2014Population}.
To better reflect how residents conceptualize the city, we also map tweets to neighborhoods as drawn by the LA Time's Mapping LA project \cite{latimesMappingLA}. The project breaks down Los Angeles County into 272 unique neighborhoods based on groupings of Census tracts modified to represent municipal boundaries and locally recognized names.

\subsection{Emotion Taxonomy}

\begin{figure*}[h]
\centering
\includegraphics[width=0.95\textwidth]{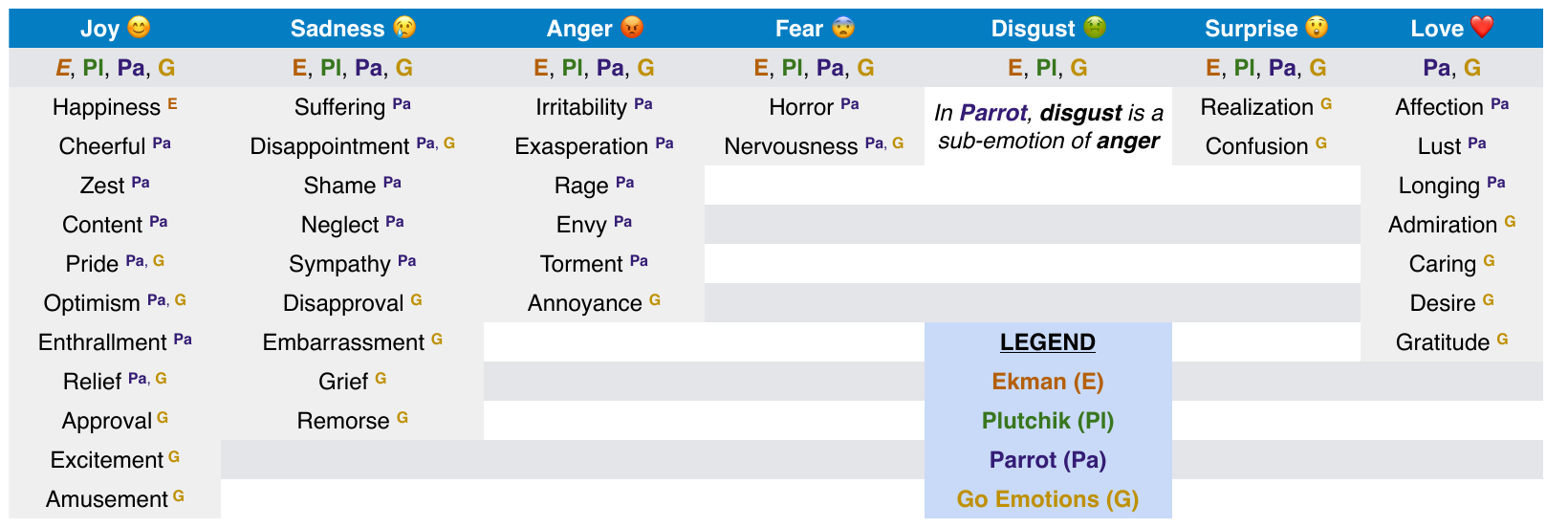}
\caption{Over forty fine-grained emotions distilled into seven primary emotion categories. %Diagram of how labels from different emotion conceptualizations are mapped to seven primary emotions. 
Each secondary emotion is labeled with emotion conceptualizations containing the emotion.
Difficulty in grouping emotions highlights inherent tensions with categorical emotion theories.}
\label{fig:emotionmap}
\end{figure*}

To support result interoperability, we distill over forty fine-grained emotions from leading categorical emotion frameworks onto seven primary categories: \textit{love, joy, surprise, anger, sadness, fear}, and \textit{disgust}, along with \textit{neutral} for non-emotional text. These primary categories are the union of Ekman and Parrott's models, with Figure \ref{fig:emotionmap} mapping Plutchik and GoEmotions taxonomies onto this core set and annotating secondary emotions by source framework. 
While this taxonomy is neither comprehensive nor fully adequate in enunciating the range of human emotion, it provides a practical framework to study affect recognition tasks.

\subsection{AAVE Dialect Density}
\label{sec:ddm}
Dialect density refers to the concentration of dialect-specific features within a given text.
Different individuals using the same dialect may prefer different features and vary dialectal presence at the sentence level \cite{chance2025re}, and different dialectal features and usage varies by region. Therefore, we estimate the strength of AAVE in a single social media post using a continuous variable called \textbf{dialect density}, which measures the concentration of AAVE-associated linguistic features. Visualized in Figure \ref{fig:ddmmap},
we draw on computational approximations of sociolinguistic features and
based on the feature set derived from \citet{Johnson2022} due to its linguistic grounding and interpretability. 
These features are approximated using NLP methods like regular expressions,  dependency parsing or named entity recognition with SpaCy \cite{spacy2025}, and language model perplexity differences when perturbing texts to GAE equivalents. %
% \katie{Whic de≠pendency toolkit did you use? How do dependency parsers do on AAVE, especially with some of the syntactic changes? Just fed a few into Stanza, and it seems to be doing pretty good, though alternate spellings seem to throw it off.}
We create a vector for feature scores, mapping each feature within range [0,1] using min-max normalization. The average over this normalized vector yields a single-valued score representing \textbf{Dialect Density Metric} (DDM).

Full implementation details of each feature approximation method used in calculating DDM is located in Appendix Table \ref{tab:ddm_feats}. 
We use 6 off-the-shelf features from \citet{Johnson2022}, expand 3 features, introduce 6 features and remove 1 feature. The off-the-shelf features include \textbf{Subject-Verb Agreement}, where third person singular presence is absent \cite{Johnson2022}. The removed feature is \textbf{Existential They}, capturing the substitution of "there" for "they". %(i.e. \chris{todo}).

Notably, natural language processing toolkits have shown poor accuracy for AAVE \citet{blodgett-etal-2018-twitter}. Therefore, we modify three inference methods to align the intended pattern with parsing behaviors of models, or to better capture nonstandard spelling and orthographic variation. Appendix Table \ref{tab:ddm_feats} gives technical details of feature changes.
\begin{itemize} 
    \item \textbf{Aint: }Expansion of regex search string \textit{ain't} to include punctuation variants (\textit{aint}) and spelling variants (\textit{yeen} \cite{Jones_2015a} which means "you ain't"). 
    % \katie{for the less-cool/less-informed folks like myself, you may need to explain what "yeen" means. I had to look it up lol}
    \item \textbf{Completive Done: }Identification of \textit{``done" + [past tense verb]} used to mark completed actions, such as "she \textit{done} left already". Dependency parsing transitioned to part-of-speech tagging to enhance accuracy \cite{stewart2014now}. 
    % \katie{imo, you might consider including an example. Also, part of speech sounds like it would be more accurate for "done + [past tense verb]"}
    \item \textbf{Invariant "Be": } Also known as \textit{Habitual "Be"}, this form denotes habitual or frequent actions and syntactically follows the subject (i.e. ``I be out every Friday" or "This be Heywood"). We expand this feature to improve accuracy \cite{stewart2014now,946e0a4a-f4f7-32aa-9aeb-bbeb8cb02903}.
    % \katie{How did you scrape these samples? Also, I might just label it as the invariant "be" as there are instances where the reading isn't habitual ("This be Heywood"), see Labov 2021 and Harris 2019 for more examples/discussion.}
\end{itemize}

\noindent 
We introduce six features to the DDM metric algorithm. 
These features are chosen because they broaden coverage of AAVE features potentially relevant to the task (e.g., profanity-based features), or various prior works that study AAVE in the Southern California region highlight these features as prominent to this regional AAVE form \cite{Rickford}.
% \katie{the separation of these two lists is a little unclear to me. Ass camouflage (new term to me, love it), copula deletion, n-use aren't more LA-specific to my knowledge than the above list, so the separation is a little odd to me.}
\begin{itemize}
    \item \textbf{Abbreviations:} Regex search for common abbreviations of ``I don't" (\textit{iont}, \textit{iono}, \textit{ioneem}) and ``Talking about" (\textit{talmbout}) based on AAVE regional patterns \cite{Jones_2015a}. Accounts for online instantiations of AAVE.
    \item \textbf{Ass Camouflage}: Special compound word used a metonymic pseudo-pronoun (i.e. “I divorced his ass.”) or an intensifier (i.e. “We was at some random-ass bar.” ) \cite{spears2013african, ziems-etal-2022-value}.
    \item \textbf{Continuative Steady:} First identified within Los Angeles \cite{baugh1979linguistic}, this feature captures use of ``steady" to denote an action occurring continuously and intensely (i.e. “She steady complaining about him.”). Approximated by ``steady" followed by a non-noun \cite{stewart2014now}. %This misses some forms of the feature.  \katie{Careful with just "non-noun" because steady can be a verb that takes a DP (e.g. "steady the boat" where the following is a determiner/article. You might consider looking for steady + progressive verb (ing verb) (with a caveat to watch out for gerunds, though I don't think those would be all that common, "it can be hard to reach steady dating" just sounds weird).} \chris{I think this is a limitation in which we generally discuss more robust approaches to this overall algorithm/metric}
    \item \textbf{Copula Deletion/Absence:} Removal of copula (e.g. are-deletion, is-deletion). This feature has strong LA presence \cite{Rickford, legum1971speech} and we capture using dependency parsing via pronoun then a noun \cite{stewart2014now} (i.e. “She a nurse.”). %\katie{There's a bit of a debate whether the copula is deleted or is there but silent, so maybe call it "copula absence". Bit curious on the specifics of how you deployed parsing for this.}
    \item \textbf{N-Use:} Regex detection for n-word \cite{spears2013african}. Not exclusive to Southern California, though potentially relevant to emotion detection due to inability to distinguish between derogatory and reclaimed usage. 
    % \katie{how would regex capture positive/negative/neural tones?}
    \item \textbf{Slang: }Regex search for miscellaneous slang terms (e.g. \textit{jawn}, \textit{finna}, \textit{cuh}) in at least 10K tweets from \cite{Jones_2015a} and popular during our dataset's timeframe. 
\end{itemize}

Dialect density provides a tractable and linguistically-grounded proxy for AAVE presence, although it does not capture the full complexity or nuance of AAVE. We emphasize that DDM is an approximation of AAVE, rather than a definitive measure of dialect identity. Further, this is not an exhaustive list of AAVE features, nor do these approximations have perfect performance. See further discussion in Limitations.

\subsubsection{Faithfulness to Region Demographics.}
To validate our DDM metric's faithfulness to Southern California AAVE, Figure \ref{fig:f1} plots the percent of each neighborhood that is African American and Figure \ref{fig:f2} shows the average dialect density of geo-coded tweets from that neighborhood. 
On observations, the two images share similar neighborhoods, including Inglewood, View Park-Windsor Hills, Gramercy Park and Baldwin Hills/Crenshaw being the most prominent. We verify this visual by running Pearson correlation between each tweet's DDM and racial composition of the author's neighborhood. We observe that Percent African American is strongly correlated with DDM (Pearson = .68, p $<$ .01) and percent White is negatively correlated (Pearson = -.66,  p $<$ .01).
On this dataset, our AAVE strength metric (DDM) is stronger in predominantly African American neighborhoods, and weaker for predominantly White neighborhoods.

\subsection{Human Annotations}

We have 12 annotators annotate a tweet sample for primary emotions as shown in Figure \ref{fig:emotionmap}. We sample 875 tweets, half from high DDM (defined as DDM $\geq$ 0.07) and half from the remaining tweets (``low" DDM). 
Annotators label each tweet for all 7 emotions where 1 denotes no emotion, 2 is slight presence, and 3 denotes strong presence. 
There are 12 total annotators, four of which are authors of this paper. 
Three annotators are African American AAVE speakers (ingroup) and nine are non-AAVE speakers (outgroup). 
We adopt a semi-ground truth approach to account for both shared perception and strong individual signals.
 Specifically, an emotion is considered present if both annotators label it as 2 (slight presence), indicating consensus around subtle expression, or if at least one annotator marks it as 3 (strong presence), indicating a salient emotional signal detectable even if not shared. This balances the need for inter-annotator agreement 
 with recognition that strong emotion may be perceived unilaterally, which is especially relevant in linguistically or culturally marked expressions where individual interpretation may vary.
Additionally, as annotator identity influences data annotations \cite{sap2021annotators}, semi-ground truth labels for high DDM tweets are determined solely by ingroup community annotators.
The full annotation set is called a ``silver" label set, saying silver rather than gold because a small community sample can not speak for the full group, particularly in a task as subjective as emotion detection.

\begin{table}[h!]
\centering
\small
\caption{Comparison of Models on Jensen-Shannon, Refusal Change, and Correlation}
\begin{tabular}{lcccc}
\toprule
\textbf{Model} &  \textbf{Jensen-Shannon} & \textbf{$\Delta$ Refusal} & \textbf{Pearson} \\
\midrule
NRC           & 0.014           & -        & 0.99 \\
RobertaGo     & 0.076           & -        & 1.00 \\
SpanEmo       & 0.059           & -        & 0.97 \\
Deepseek-Qwen & 0.045 (±0.020)  & -0.002   & 1.00 \\
Llama-3.1     & 0.025 (±0.007)  & -0.000   & 1.00 \\
GPT-4o-Mini        & 0.011 (±0.001)  & -0.000   & 1.00 \\
Latimer-4o-Mini       & 0.013 (±0.002)  & -0.001   & 0.99 \\
\bottomrule
\end{tabular}
\caption{Similarity between model behavior on sample and overall dataset. Small changes in prediction frequencies and refusal rates indicate a representative sample.}
\label{tab:sample_rep}
\end{table}

To ensure reliable analyses of model predictions, we validate how representative the small sample in comparison to the full 2.7M dataset and report results in Table \ref{tab:sample_rep}. 
We compute Pearson's correlation between vectors of emotion prediction frequencies, finding correlations are extremely strong, always averaging at least 0.97 and individually never dropping below 0.90.
We measure Jensen-Shannon Divergence between emotion label frequencies of the sample and full distributions, where values towards 1 indicate lower similarity, and find all values to be extremely low ($\leq$ 0.1). 
Finally, $\Delta$ Refusal measures the change in refusal rate between overall and sampled distributions. Entries appearing 0 but with + or - symbols indicate minuscule change. Blank values indicative of non-generative models being unable to refuse instances. Absolute difference between refusal rates on sample and overall are never larger than 0.002\%. Model prediction distributions on the annotated sample are similar to the distributions of the overall dataset.

\subsection{Model Choice}

\subsubsection{Classification Models.} We compile emotion classification models from popular computational social science venues. Specifically, we extract 2023 and 2024 papers from ICWSM, WebSci, SBP-BRiMS and WWW containing the string \textit{emotion} in the title. 
We find 21 papers: 2 use LIWC and 2 NRC lexicons, 3 use SpanEmo \cite{chochlakis2023using}, 3 BERT-based models, 4 introduce fine-tuned LLMs, 4 with unspecified emotion modeling techniques and 5 papers not containing emotion modeling at all.
Accordingly, we use lexicon \texttt{NRC} (Plutchik-based), 
and BERT-based models \texttt{SpanEmo} (Plutchik $\cup$ Ekman) and \texttt{roberta-go-emotions} (Go Emotions).
NRC and SpanEmo return soft scores representing likelihood of a text containing each emotion, while roberta-go-emotions (RobGo) uses a distribution over 20+ emotions. We consider an emotion to be present if its score is 0.05 or above.

\subsubsection{Generative Models and Prompting.}
For text generation models, we use state-of-the-art models \texttt{GPT-4o-mini}\footnote{https://platform.openai.com/docs/models/gpt-4o-mini} \cite{hurst2024gpt}, \texttt{Deepseek-R1-Distill-Qwen-7B}\footnote{https://huggingface.co/deepseek-ai/DeepSeek-R1-Distill-Qwen-7B} \cite{guo2025deepseek} and \texttt{Llama-3.1}\footnote{https://huggingface.co/meta-llama/Llama-3.1-8B} \cite{dubey2024llama}. 
To investigate the benefit of finetuning on diverse and linguistically-inclusive set of texts, we additionally include \texttt{Latimer}, a fairness-oriented model built upon GPT and fine-tuned on ``indigenous folk tales, community-driven oral histories, and grassroots publications from various parts of the world" \cite{LatimerAI}.
 Due to resource limitations, GPT and Latimer are run on a sample of 2500 samples rather than all 2.7M tweets. 

\subsubsection{Prompt Design and Evaluation.} 
\begin{figure}[h]
\centering
\includegraphics[width=.8\textwidth]{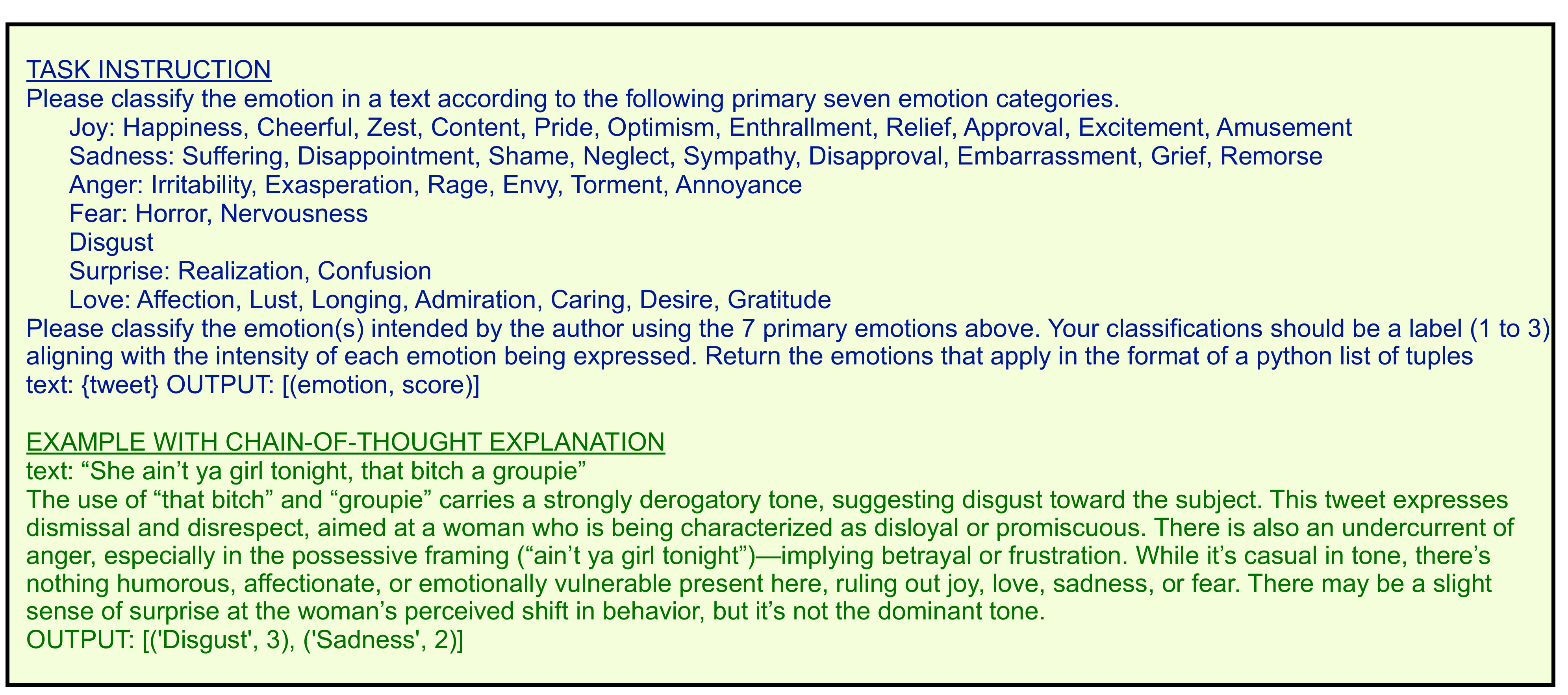}
\caption{Example Prompting Schema. Blue text indicates zero-shot (\texttt{zero}) prompting schema. Few-shot (\texttt{few}) expands to include three (text, output) pairs as shown in purple. Finally, chain-of-thought (\texttt{COT}) prompting schema adds reasoning steps as featured in green.}
\label{fig:promptexample}
\end{figure}

We employ three prompt configurations that vary in number of provided examples and example explanations. The zero-shot (\texttt{zero}) explanation consists of a task explanation and output format specification. The task instruction asks the LLM to label a text with one or more emotions using our distilled emotion conceptualization (Figure \ref{fig:emotionmap}). The few-shot (\texttt{few}) configuration provides the model 3 labeled examples of expected labels. The chain-of-thought (\texttt{COT}) schema adds step-by-step explanations to each labeled example \cite{wei2022chain}. Figure \ref{fig:promptexample} displays an example of schema components.

As models are highly sensitive to the phrasing of user input, we design 6 task instructions to be randomly chosen from at test-time.
We manually write 20 prompts and use the 5 highest-performing on Llama-3.1 were selected. Ideally we would check performance against all models, however due to limited resources we analyze just one. 
These prompts are designed to ask for a range of output formats and include some perturbations linked to higher performance (e.g. ``I will tip you \$10"), as accuracy of tasks may be altered by the smallest prompt perturbations \cite{salinas2024butterfly}.
For the \texttt{cot} prompting schema, we choose examples to annotate based on instances misunderstood by non-AAVE speaking authors. Explanations are written by AAVE-speaking authors.
Models are run with temperature set to 0.70.

\section{Results}

\subsection{RQ1: How does group membership influence emotion annotations of AAVE and GAE texts?}

\begin{figure}[h!]
\centering
\includegraphics[width=.7\textwidth]{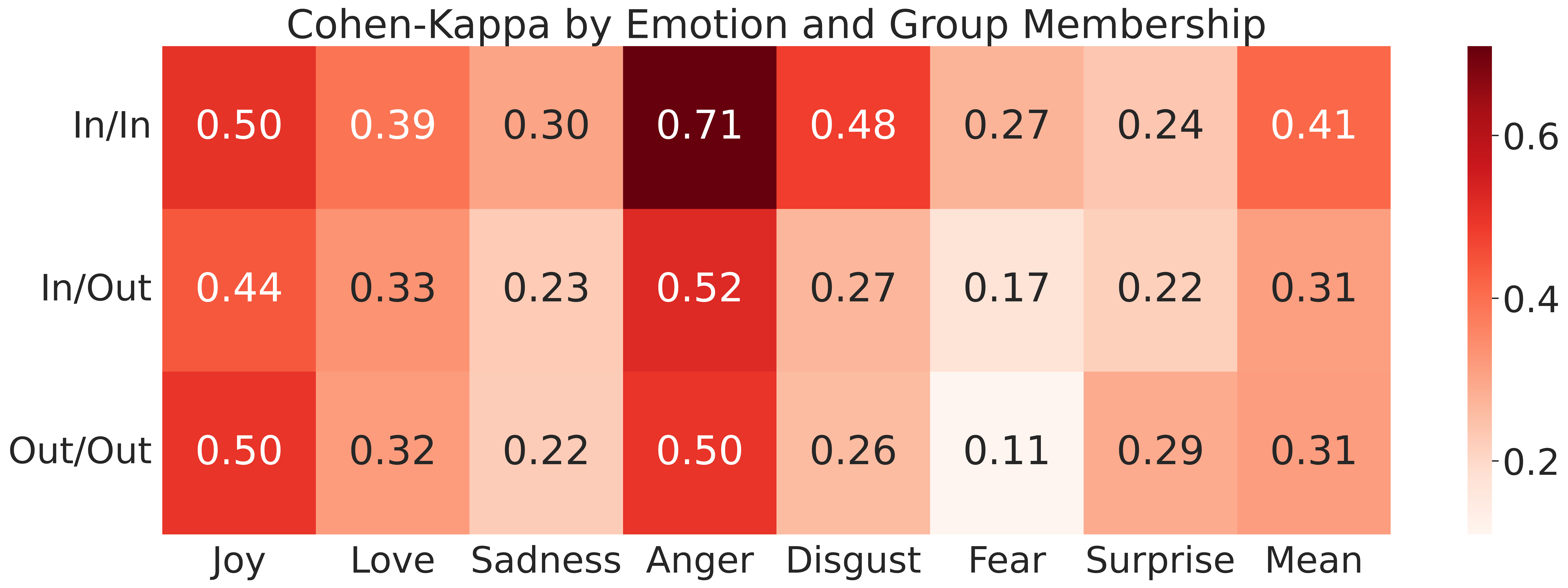}
\caption{Annotator agreement measured by Cohen's Kappa averaged over pairs of annotators, stratified by group membership and individual emotion. Y-label denotes annotator subset of agreement calculation, where "In/In" is calculated between ingroup members, "In/Out" between ingroup and outgroup member pairings, and "Out/Out" between outgroup annotators. Agreement is highest for joy and anger, particularly within ingroup annotations.}
\label{fig:ck_human}
\end{figure}

\subsubsection{Annotator Agreement.}
Cohen's Kappa ($\kappa$) measures frequency that two annotators prescribe the same labels to the same text while accounting for chance agreement. 
We look at Cohen's $\kappa$ both overall and fine-grained by annotator group membership, taking mean of pair-wise agreement values.  Figure \ref{fig:ck_human} shows low inter-annotator agreement across all annotators (Cohen Kappa $\kappa$ = 0.32). We observe that ingroup 
agreement is highest for anger ($\kappa$ = 0.71), joy ($\kappa$ = 0.50), and disgust ($\kappa$ = 0.48). 
Outgroup agreement is highest for joy and anger ($\kappa$ = 0.50) followed by love ($\kappa$ = 0.32). 
Looking at $\Delta\kappa$, the difference in agreement rate, agreement between ingroup and outgroup is often higher than outgroup but lower than ingroup, with the highest discrepancies on surprise ($\Delta\kappa$ = 0.07) followed by a tie between joy and fear ($\Delta\kappa$ = 0.06). 
Anger and joy appear easier for annotators to agree upon, while fear is more difficult.

\begin{figure*}[h!]
\centering
\includegraphics[width=\textwidth]{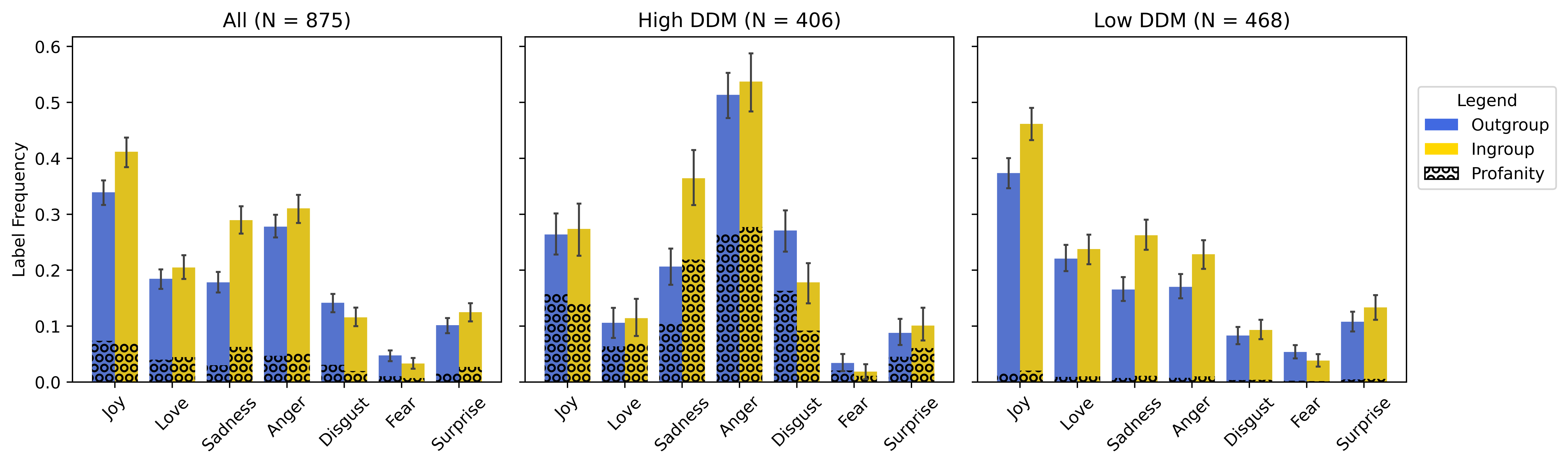}
\caption{Emotion label frequency in annotated sample. Colors denote ``ingroup" membership. Two DDM features are profanity-based, hence the hatching pattern marks proportion of tweets containing profane language.}
\label{fig:annot_freq}
\end{figure*}

\subsubsection{Differences by Group.}
The role of community membership in annotations is analyzed by looking at emotion label distributions. Shown in Figure \ref{fig:annot_freq}, ingroup annotators identify more emotion, regardless of dialect density, except for disgust, which outgroup members see more of in high DDM tweets. We run a one-way ANOVA test, where for each emotion we compare binary predictions from ingroup versus outgroup annotators. Results on the whole annotated set find significant differences for sadness (F = 18.8, p < 0.01), anger (F = 5.8, p < 0.01), fear (F = 6.2, p < 0.01) and disgust (F = 4.6, p < 0.05). Repeating the test on only high DDM instances, these same emotions are different by community membership (all p < 0.05). The test is repeated only on tweets containing profanities, and find a different set of emotions to vary by membership: sadness (F = 31.1), disgust (F = 16.7) and love (F = 5.2) (all p < 0.05). Community membership appears linked to differences in emotion annotations.

\subsubsection{Silver Label Set}
Computing model performance necessitates some approximation of ground truth. For the annotated sample of 875 instances,
``silver" labels are the mode (i.e., most frequent) label from a set of annotators.
If multiple modes exist, we instead use the average of the top labels. 
When the most frequent labels are on opposing extremes (e.g., `No emotion presence' vs. `High emotion presence'), disagreements are mapped to `Slight presence', or rounded up to 1 when interpreting results as a binary classification problem.
Of these, 406 tweets have high DDM (DDM $\geq$ 0.07), with 
76\% of instances annotated by 2 people and 24\% labeled by 1. ``Silver" labels here are determined solely by ingroup annotators. We observe 42 of these tweets feature extreme disagreement: joy (8), love (2) , sadness (26), anger (4), disgust (3), fear (1), and surprise (3). 
The remaining 468 tweets exhibit lower DDM (DDM $<$ 0.07) and are labeled by all 12 annotators. Consequently, the remaining data has substantially more annotators per instance (mean annotators/post = 3.4, $\sigma$ = 1.1). There are 43 posts with extreme disagreement (48 emotion markers): joy (6), love (19), sadness (4), anger (9), disgust (6), fear (3), and surprise (1). Silver labels are determined using community-informed majority vote.

\subsection{RQ2: How do emotion AI models perform on AAVE text compared to GAE text? Which annotation groups do model predictions align with?}

\subsubsection{Performance on the Whole Dataset. }

\begin{figure}[h!]
\centering
\includegraphics[width=.8\textwidth]{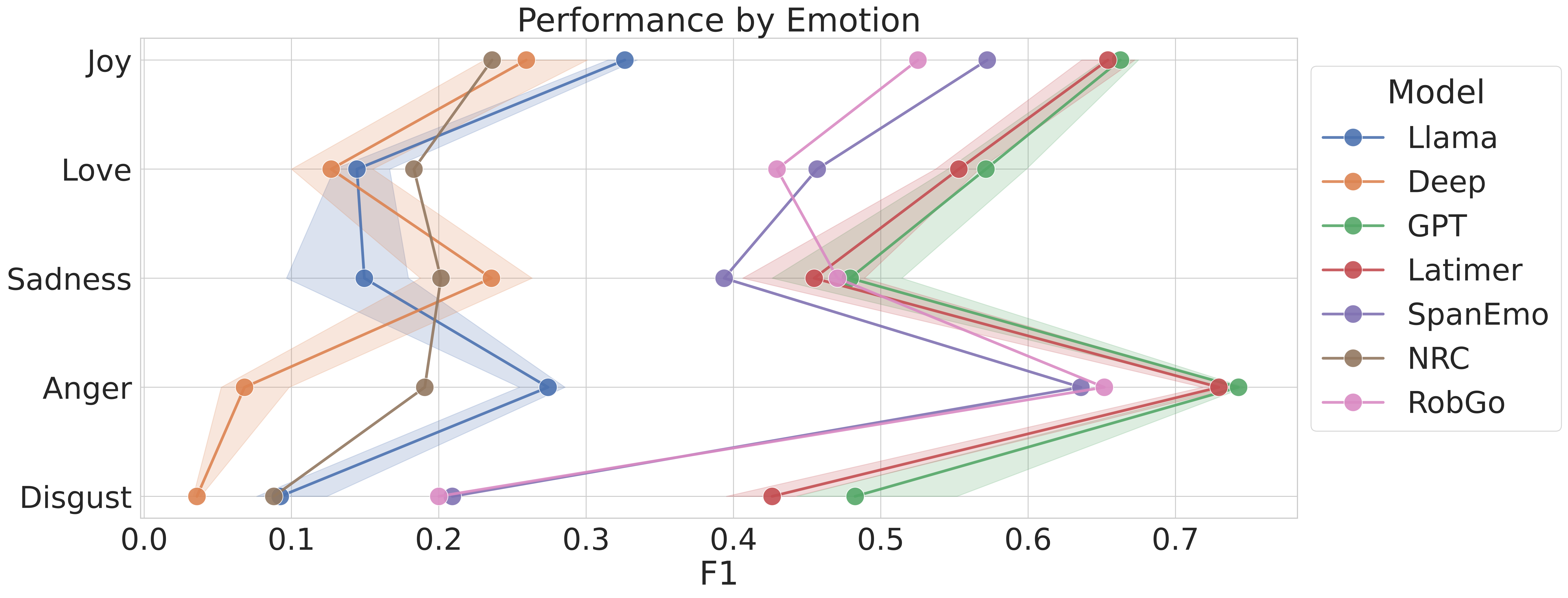}
\caption{Model F1 for each emotion. The GPT-based and BERT-based models are more right-leaning, indicating higher performance. Emotions with less than 60 positive labels excluded.}
\label{fig:F1}
\end{figure}

Performance is evaluated using precision, the fraction of positive predictions that are correct, and recall, the fraction of positive instances correctly identified. Full precision and recall results by model and emotion are reported in Appendix Table \ref{tab:PR}. To facilitate comparison, Figure \ref{fig:F1} presents average precision and recall across emotions with at least 60 positive annotations.

Averaging over all emotions, GPT-4o-Mini achieves the highest overall performance ($\mu$(P) = 0.55, $\mu$(R) = 0.70), with Latimer-4o-Mini close behind ($\mu$(P) = 0.52, $\mu$(R) = 0.67). The model with the lowest precision and recall is Deepseek-Qwen ($\mu$(P) = 0.19, $\mu$(R) = 0.14) followed by Llama-3.1 ($\mu$(P) = 0.19, $\mu$(R) = 0.21). 
Overall, emotion performance is poor to moderate with BERT-based and GPT-based models performing strongest. These well-performing models tend to obtain recall scores higher than precision. We observe that the discriminative applications of BERT outperform generative models Deepseek-Qwen and Llama-3.1.

Overall performance is measured with F1 score, defined as the harmonic mean of precision and recall, visualized by model and emotion in Figure \ref{tab:PR}. Emotions are loosely ordered by valence, with joy and love representing positive affect and sadness, anger, and disgust representing negative affect. For generative models, within-model variance across prompting schemas is highlighted.
The models with the highest overall F1 are GPT ($\mu$(F1) = 0.59), Latimer ($\mu$(F1) = 0.56), Roberta Go ($\mu$(F1) = 0.46) and SpanEmo ($\mu$(F1) = 0.45). Poorer performance is observed from Deepseek ($\mu$(F1) = 0.15), NRC Lexicon ($\mu$(F1) = 0.18) and Llama-3.1 ($\mu$(F1) = 0.20). These groups appear linearly separable across individual emotions, indicating consistent stability in cross-emotion rankings.

Model performance varies substantially across emotions. High-performing models (GPT, Latimer, RobertaGo, and SpanEmo) achieve their best F1 on anger, while lower-performing models peak on joy. All models exhibit their weakest performance on disgust, regardless of architecture. Performance suggests that lexical and contextual cues for anger and joy are more readily captured in existing emotion representations, whereas subtler or more context-dependent emotions such as disgust remain challenging for current emotion AI systems.

We assess the effect of prompt schema (zero-shot, few-shot, and chain-of-thought)  on generative model performance. Differences in F1 reflect how model accuracy changes with additional context. Across all generative models, changes are minimal ($\Delta F1 \leq$ 0.007). Few-shot prompting yields a slight improvement over zero-shot for Deepseek-Llama and GPT-4o-Mini, whereas CoT prompting marginally decreases performance for all four models. The largest decline occurs for Deepseek-Llama, with a 1.2-percentage-point reduction in mean F1. Overall, prompting schema has negligible influence on emotion-recognition performance.

\subsubsection{Similarity within GPT Family}
Comparing GPT-4o-Mini and Latimer-4o-Mini, model predictions converge with increasing contextual information. Using cosine distance between full prediction vectors (excluding refusals), zero-shot predictions are most divergent (0.33), while few-shot (0.14) and CoT (0.16) predictions are substantially closer. Additional prompting context thus appears to align model outputs. One possible explanation is that chain-of-thought reasoning encourages models to rely more heavily on internal priors than on user-provided input, reducing diversity in model behavior \cite{chochlakis2024strong}.

\subsubsection{Sensitivity to Prompt Perturbations}
We look at the change in performance with task instruction details like output format and specific task wording. After testing 6 prompts, 5 prompts result in similar model performance to the model's average F1 ($\Delta$F1 $\leq$ 0.03) with 1 outlier yielding worse performance ($\Delta$F1 = 0.08).
The poor-performing prompt uniquely formats emotion categories using a manually-formatted arrow (e.g. \textit{surprise $\rightarrow$ realization, confusion}) and does not use any performance-enhancing keywords like please and thank you.

\subsubsection{Model Performance by Dialect Density.} 
\paragraph{Precision and Recall Disparity.} 
Change in precision and recall on high-DDM instances is reported in the rightmost column of Table \ref{tab:apr_comp}. 
We report the difference on the high DDM sample, as the low DDM sample inherently shows the opposite trend. 
We observe that models NRC, Deepseek-Qwen and Llama-3.1 all have mean precision rates at least 5\% higher on AAVE tweets compared to the whole dataset. These three models are also the low-performing models. Model discrepancies in average recall never exceed a difference of 3\%. Mean precision increases with high DDM for low-performing models. 
Precision and recall in aggregate show a small change in precision on poor-performing models, necessitating analyses disaggregated by emotion and error type.

\begin{table}[h!]
\centering
\begin{tabular}{l
                S[table-format=1.3] S[table-format=1.3]
                S[table-format=1.3] S[table-format=1.3]}
\toprule
\multirow{2}{*}{\textbf{Family}} & \multicolumn{2}{c}{\textbf{All}} & \multicolumn{2}{c}{\textbf{High DDM}} \\
\cmidrule(lr){2-3} \cmidrule(lr){4-5}
 & {Precision} & {Recall} & {Precision} & {Recall} \\
\midrule
NRC     & 0.20 & 0.18 & 0.28 {\textcolor{green!50!black}{$\uparrow$}} & 0.21 \\
RobGo   & 0.45 & 0.52 & 0.44 & 0.50 \\
SpanEmo    & 0.33 & 0.82 & 0.35 & 0.79 \\
Deepseek-Qwen    & 0.19 & 0.14 & 0.26 {\textcolor{green!50!black}{$\uparrow$}} & 0.15 \\
Llama-3.1   & 0.19 & 0.21 & 0.24 {\textcolor{green!50!black}{$\uparrow$}} & 0.21 \\
GPT-4o-Mini     & 0.55 & 0.70 & 0.57 & 0.69 \\
Latimer-4o-Mini & 0.52 & 0.67 & 0.55 & 0.66 \\
\bottomrule
\end{tabular}
\caption{Precision and recall averaged per model family. Performance is reported for the labeled sample and a subset of high dialect density instances. Colored arrows denote whether the performance metric is higher or lower than the overall metric. At a bird's-eye view, the lower-performing models appear to be more precise on high DDM instances.}
\label{tab:apr_comp}
\end{table}

\begin{figure*}
    \centering
    \includegraphics[width=\textwidth]{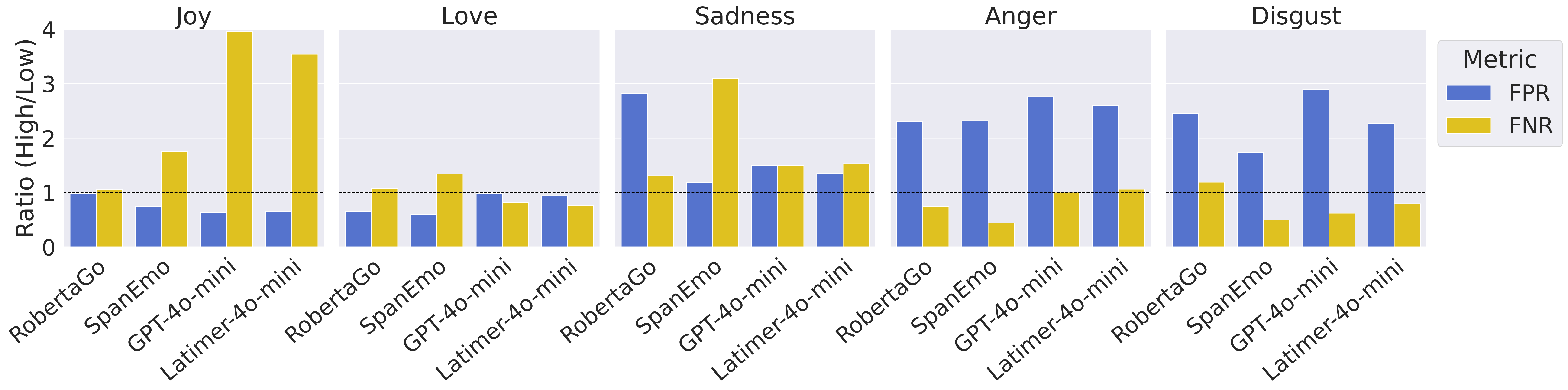}
    \caption{Ratio of false prediction rates from high to low DDM instances. Performance reported for the four strongest models. GPT-based systems exhibit false negative predictions of Joy much higher on AAVE-dense texts. All high-performing models falsely predict anger and disgust in AAVE-dense texts higher than in GAE texts.}
    \label{fig:fpr_disp}
\end{figure*}

\paragraph{Misclassification Rates.} 
Misclassifications occur when models predict emotion presence when there is none (false positive) or models predict no emotion when there is presence (false negative).
We investigate False Positive Rate (FPR), the frequency of false positives, and False Negative Rate (FNR), the frequency that models falsely predict no emotion presence.
For each metric, we plot the ratio of FPR or FNR on high DDM (AAVE) instances over the metric on low DDM instances. Figure \ref{fig:fpr_disp} shows  FPR ratio disparity ($\Delta$FPR  = FPR$_{AAVE}$/FPR$_{GAE}$) in blue, and FNR ($\Delta$FNR  = FNR$_{AAVE}$/FNR$_{GAE}$) ratio disparity in yellow.
The plot reports disparity rates for the high-performing BERT and GPT-based models on emotions with at least 60 positive annotations.

In our sample set, models are more likely to falsely interpret anger and disgust in texts with AAVE features.
The green bars show $\Delta$FPR on the annotated sample using the silver labels.
All plotted models falsely predict anger on AAVE at rates more than double GAE, with GPT leading the disparity ($\Delta$FPR = 2.8, $\mu$(FPR) = 0.12), followed by Latimer ($\Delta$FPR = 2.6, $\mu$(FPR) = 0.11), then tied between RobGo ($\Delta$FPR = 2.3, $\mu$(FPR) = 0.15) and SpanEmo ($\Delta$FPR = 2.3, $\mu$(FPR) = 0.37). SpanEmo has the lowest mean false positive rate, meaning even though it has the ``lowest" ratio disparity, it has the largest FPR difference: on low dialect density samples, SpanEmo's FPR is 25\%, while on high dialect instances the FPR climbs to 60\%. 
Aside from SpanEmo, disgust increases in disparity for all models: GPT ($\Delta$FPR = 2.9, $\mu$(FPR) = 0.15), RobGo (2.5, , $\mu$(FPR) = 0.02) and Latimer (2.3, $\mu$(FPR) = 0.16).
Emotions with positive valence result in $\Delta$FPR much lower than 1, meaning false positives are more likely on low-AAVE (i.e., GAE) instances. 
Better understood using the opposite ratio ($\Delta$ FPR$^{-1}$ = FNR$_{GAE}$/FNR$_{AAVE}$), we find strong disparities on false positives for joy for GPT ($\Delta$FPR$^{-1}$ = 1.6, $\mu$(FPR) = 0.35), Latimer (1.5, $\mu$(FPR) = 0.39) and SpanEmo (1.4, $\mu$(FPR) = 0.54).
In the AAVE texts in our sample, the highest-performing models predicted more false positives for anger and disgust, and fewer false positives for joy.

False negative rates in emotion detection highlight which emotions models miss. Joy exhibits one of the most striking disparities, with GPT having a false negative rate 4x higher on AAVE texts ($\Delta$FNR = 4.00, $\mu$(FNR) = 0.09), with additional strong disparities for Latimer (3.5, $\mu$(FNR) = 0.08) and SpanEmo (1.8, $\mu$(FNR) = 0.11). 
For sadness, SpanEmo exhibits uniquely high ($\Delta$FNR = 3.1, $\mu$(FNR) = 0.32) with other models showing moderate disparity (1.3 $\leq$ $\Delta$FPR  $\leq$ 1.5).
False negative rate disparities reveal GPT-4o-mini and Latimer-4o-mini to miss joy in AAVE texts at rates 3.5 to 4 times higher than in GAE texts.

\paragraph{Bias Mitigation by Latimer.}
To assess the impact of fine‑tuning on diverse, community‑driven texts, we compare Latimer against GPT‑4o‑mini in $\Delta$FPR and $\Delta$FNR.  Latimer reduces FPR disparity on disgust by 6 precision points and anger by 2 points. The FNR disparity on joy is reduced by 2 points.  These results demonstrate that inclusive, dialect‑aware fine‑tuning improves model fairness on dialect‑rich text, without introducing major errors on dialect‑sparse inputs. However, more work is needed to understand the role of model priors coming out in response to elongated, chain-of-thought reasoning.

\begin{figure}[h!]
\centering
\includegraphics[width=\textwidth]{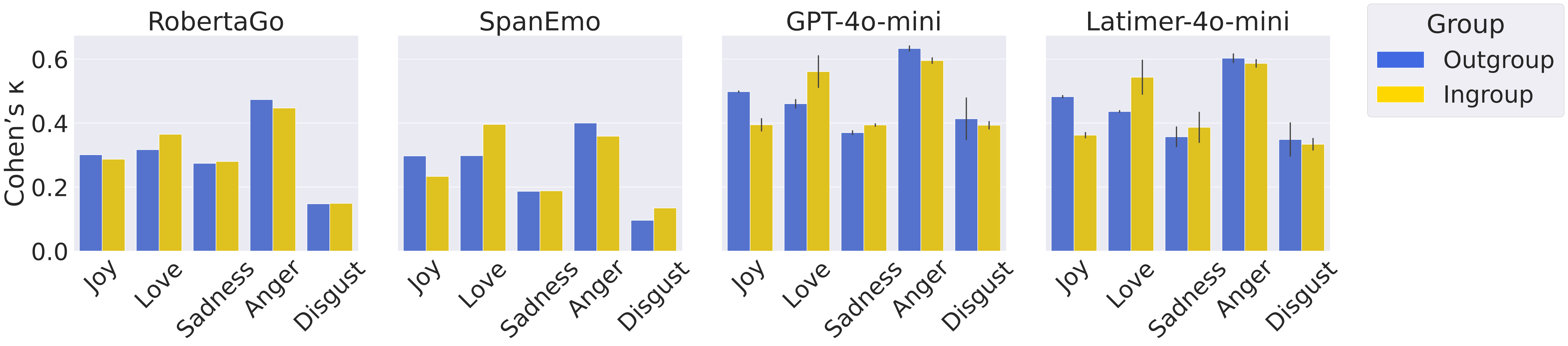}
\caption{Model agreement with human annotations via average pairwise Cohen's Kappa. Models have higher agreement with outgroup annotations for joy, anger, and disgust.}
\label{fig:ck_model_human}
\end{figure}

\subsubsection{Model Prediction Similarity by Group.}
Model alignment with annotation groups is measured using treating models as annotators and computing pair-wise Cohen's Kappa annotator agreement.
The three low-performing models all yield Cohen's Kappa values no greater than 0.10, and often with negative values, indicating an agreement so poor that random choice is more likely to agree with human annotations than with model predictions.
We plot agreement by emotion on high-performing models in Figure \ref{fig:ck_model_human}. 
Top agreement with both ingroup and outgroup human annotators is achieved by GPT-4o-Mini and Latimer-4o-Mini, though all four models typically have higher agreement with outgroup labels. Notably, love consistently shows higher agreement with ingroup labels.
We observe trends between model agreement and emotion-level precision and recall rates.
RobertaGo achieves its highest agreement with both ingroup and outgroup annotators on anger, which is also the emotion with the highest precision and recall. Conversely, RobertaGo's lowest agreement for both groups is on disgust, which also yields the model's lowest emotion-level recall. 
Similarly, SpanEmo obtains its highest outgroup agreement on anger, aligning with the model's highest emotion-level precision.
SpanEmo achieves its highest ingroup annotator agreement on love, aligning with SpanEmo’s uniquely high performance on love (Recall = 0.69).
For both GPT-4o-mini and Latimer-4o-mini, anger achieves the highest agreement across membership groups. Anger is also both model's highest achieved precision. The emotion with the lowest agreement with ingroup annotations is sadness, which also yields the lowest recall performance for both models. 
High precision appears to be linked to higher agreement rates, and low recall appears to be linked to low agreement. 
Except for love, models tend to have higher agreement with outgroup annotators.

\subsection{RQ3: Which linguistic features of AAVE correlate with model performance?}

\subsubsection{Individual AAVE Feature Influence}
To understand the sensitivity of emotion labels to specific AAVE dialect features, we run a series of linear regressions on label generations.  
Specifically, we formulate each linear regression model as follows:
\begin{equation}
    y_i(x) = \epsilon_i + \beta_0 + \sum_{j \in \mathbb{N}} \beta_j
\end{equation}
Where N is the selected set of dialect density features. As 3 features are based on perplexity of GPT-2, we remove perplexity-based features from the selected feature set N. We additionally add a disturbance term $\epsilon$ to help prevent linear dependence between predictors. Predictor $y_i(x)$ represents labels of emotion $x$ from annotator $i$, where annotator $i$ may be either a model or a human. A model is learned for each emotion on each of the 4 top-performing models and each of the 12 annotators.

\begin{figure*}[!htp]
  \centering
  \includegraphics[width=\textwidth]{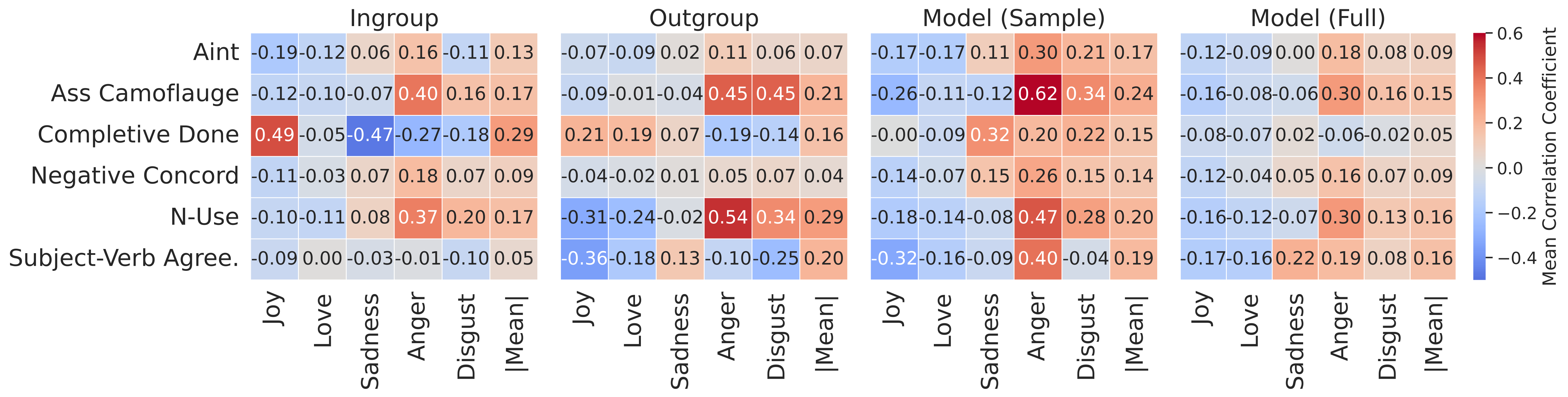}
\caption{Linear regression results. Mean correlation coefficient (p < 0.05) for individual AAVE features and labels for ingroup, outgroup, and model predictions. On our DDM-balanced sample, outgroup and model predictions are more correlated with profanity-based AAVE features. %\chris{maybe we have this be a 2 x 2 so that its more readable?}
}
\label{fig:linreg_corr}
\end{figure*}

\begin{table}[h!]
\small
\centering
\begin{tabular}{lcccccc}
\hline
\textbf{Emotion} & \textbf{Ingroup} & \textbf{Outgroup} & \textbf{SpanEmo} & \textbf{RobertaGo} & \textbf{GPT-4o-Mini} & \textbf{Latimer-4o-Mini} \\
\hline
Joy     & 0.08 $\pm$ 0.06 & 0.12 $\pm$ 0.05 & 0.10 & 0.03 & 0.05 $\pm$ 0.01 & 0.06 $\pm$ 0.00 \\
Love    & 0.05 $\pm$ 0.02 & 0.12 $\pm$ 0.10 & 0.07 & 0.06 & 0.03 $\pm$ 0.01 & 0.02 $\pm$ 0.01 \\
Sadness & 0.04 $\pm$ 0.02 & 0.09 $\pm$ 0.04 & 0.06 & 0.10 & 0.03 $\pm$ 0.01 & 0.04 $\pm$ 0.00 \\
Anger   & 0.20 $\pm$ 0.12 & 0.26 $\pm$ 0.11 & 0.27 & 0.18 & 0.17 $\pm$ 0.04 & 0.15 $\pm$ 0.03 \\
Disgust & 0.12 $\pm$ 0.14 & 0.17 $\pm$ 0.12 & 0.18 & 0.03 & 0.07 $\pm$ 0.03 & 0.06 $\pm$ 0.01 \\
\hline
\end{tabular}
\caption{Emotion-specific R$^{2}$ values from linear regression, with (mean $\pm$ standard deviation) formatting where multiple values are taken over multiple annotators or prompting schemas. Human and model annotators have the highest R$^{2}$ values for anger. }
\label{tab:rsquared}
\end{table}

Heatmap in Figure \ref{fig:linreg_corr} plots Matrix M, where M[x,y] is the mean correlation of feature $y$ when predicting emotion $x$, with corresponding R-squared values reported in Table \ref{tab:rsquared}.
Only statistically significant correlations (p < 0.05) are included in mean calculations. Right-most columns show the overall influence of each feature by plotting the mean value of the coefficients of feature y. For visual simplicity, we only include DDM features where |Mean| for model's is at least 0.10. 

The overall feature influence is approximated using |Mean|, which represents the average absolute value of statistically significant coefficients. For ingroup annotators, the most influential is \textit{completive done} (0.29), followed by a tie between \textit{ass-camoflauge} (0.17) and \textit{n-use} (0.17). For outgroup, the strongest factor is \textit{n-use} (0.29), then \textit{ass-camoflauge} (0.21) and \textit{subject-verb agreement} (0.20). 
Model predictions on the same sample are most correlated with \textit{ass-camoflauge} (0.24), then \textit{n-use} (0.20) and \textit{subject-verb agreement} (0.19). Model predictions on the whole dataset see reductions in |Mean|, likely due to high DDM instances being less than 14.8\% of the dataset, as opposed to the intentional 50\% in the labeled sample. Non-ingroup annotations appear more influenced by the presence of profanities. Model results align with previous works, finding that unless explicit guidance is provided, language models make predictions based on the presence of pejoratives, regardless of the community's relationship to the term itself \cite{dorn2024harmful}.

Comparing the correlation of AAVE features with ingroup labels versus outgroup annotations reveals model alignment behavior.
Changes in emotion columns show that outgroup annotations are more strongly correlated with anger and disgust for two-thirds of the features. Joy is less correlated with DDM for \textit{completive done}, \textit{n-use}, and \textit{subject-verb agreement}. Love is less correlated with DDM on 1/2 of the features. Sadness is correlated with both positive and negative changes, although notably \textit{completive done} shows a significant shift (-0.47 to 0.07). Looking at model prediction correlations, we see that anger and disgust are more correlated with DDM features than ingroup annotations across \textit{all} features. Joy is less correlated with DDM for most emotions, and love is less correlated for all emotions (though \textit{ass-camoflauge} is only a 0.01 difference). Sadness is less correlated with DDM for \textit{ass-camoflauge}, \textit{n-use}, and \textit{subject-verb agreement}. 
Examining the changes in outgroup and model annotations in relation to ingroup annotations, the left side appears more blue (AAVE is less correlated with positive emotions), and the right side becomes more pigmented (AAVE is more correlated with anger and disgust).

\subsection{RQ4: Incorporating Census demographic information, do some emotions show spurious correlations with neighborhood racial composition?}
\begin{figure*}[h!]
\centering
\includegraphics[width=\textwidth]{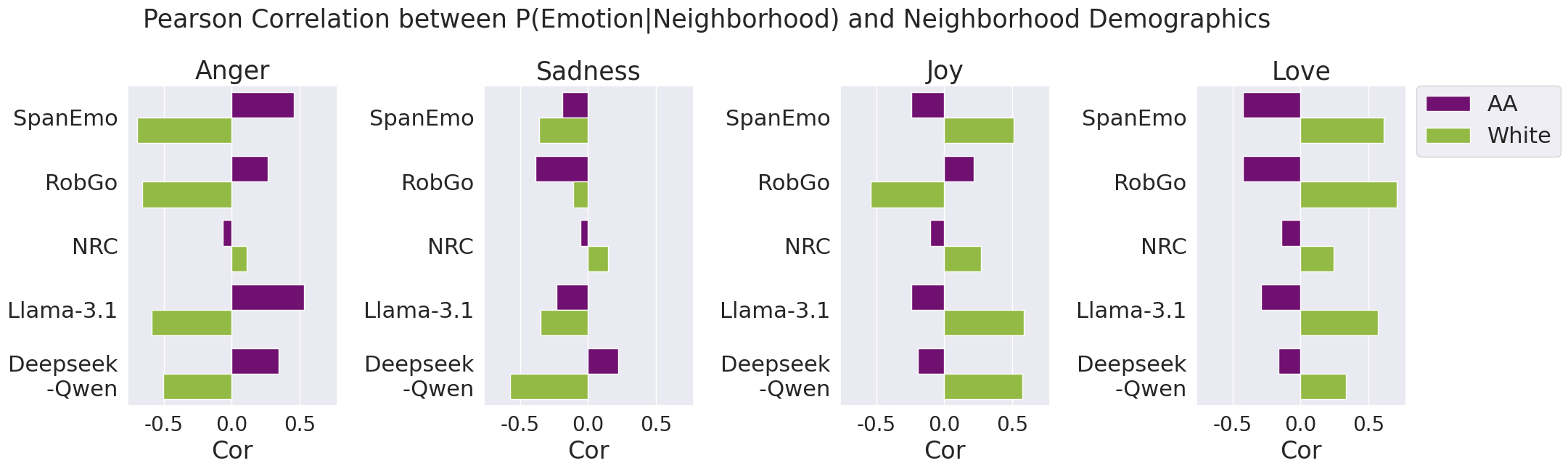}
\caption{Correlations between neighborhood-level model predictions and racial composition. GPT-4o-Mini and Latimer excluded due to low p-values, likely from small sample size (N = 2500). Anger is positively correlated with the neighborhood's percent African American and negatively correlated with the percent White. Love shows a converse association.}
\label{fig:cor_aa}
\end{figure*}

Linking back to the demographic data of tweet locations, Figure \ref{fig:cor_aa} shows Pearson correlation between each neighborhood's probability of emotion P(emotion$|$neighborhood, model) and the percentage of African American residents in that neighborhood. GPT and Latimer are excluded from the analysis due to high p-values resulting from small sample sizes, which are a consequence of the costs associated with labeling data. 
Anger displays the strongest demographic correlation, with SpanEmo, RobertaGo, Llama-3.1 and Deepseek-Llama all predicting anger in proportion to African American presence and inversely with White population. The neighborhoods with the highest predicted anger (e.g. View Park-Windsor Hills, Leimert Park) also have some of the highest African American populations. 

Positive emotions show the opposite trend. P(Joy) and P(Love) are positively correlated with White population and negatively correlated with African American population in most models. Sadness exhibits mixed associations, although it is often negative for both demographic groups.
Aggregated by median, African American demographics are positively correlated with anger (Pearson's correlation $r = 0.27$) and negatively with joy ($r = -0.10$), while White presence is positively correlated with joy ($r = 0.27$) and negatively with anger ($r = -0.51$).
Our results suggest that care should be taken when drawing conclusions about subjective well-being~\cite{jaidka2020estimating} or mental health status~\cite{Han2022} from social media data.

\section{Discussion}
This work investigates how text-based emotion detection AI systems handle African American Vernacular English. Garnering annotations on a sample of 875 tweets sourced from the greater Los Angeles County, we find that African American annotators identify more emotional expression in text compared to outgroup annotators, except for outgroup labeling excessive amounts of Disgust on AAVE-written texts. We find that models label texts with AAVE features 
with disproportionately high false positive rates for anger and disgust, while struggling to identify joy. Additionally, model predictions of anger and disgust strongly correlate with profanity-based features of AAVE. In our Los Angeles Twitter dataset, we see predictions of anger increase with African American population density and decrease with White population density. The inverse pattern holds for joy and love.

Two unique African American communication practices may partially explain the differences in annotators' emotion perception and sensitivity to profanities. In particular, \textit{augmentation}, where sounds or words are added to some phrase or clause to increase entertainment or interest, and \textit{performativity}, the self-dramatization for an audience \cite{spears2007african}. These specific practices incorporate candor or conflict, often seen as inappropriate attention-grabbing or downright obscene by White audiences (e.g., "Get out of bed" could become "Get your lazy behind out of bed" \cite{spears2007african}). In our results, African American annotators identified more emotion expression than outgroup annotators, potentially reflecting that African American annotators are more attuned to entertainment-enhancing perturbations, or that outgroup annotators miss emotion expression in more ambiguous texts. Our results additionally found that models and outgroup annotators are heavily influenced by the presence of profanities when assigning labels. Dominant linguistic norms tend to equate the presence of profanities with obscenity or offense, regardless of context \cite{pullum2018slurs}, potentially leading non-AAVE fluent annotators to read augmented texts as inappropriate or obscene.

Models label AAVE-dense texts with disproportionately high false positive rates for anger and disgust, while struggling to identify joy. The focus on anger falls in line with the "angry black person" (especially women) stereotype, which falsely posits African Americans as overly angry and hostile \cite{Wingfield_2010, walley2009debunking}. The stereotype becomes codified through frequent instances of ignoring African American emotional expression unless it is overtly angry \cite{walley2009debunking}, and it has led to White individuals disproportionately interpreting neutral or ambiguous emotional expressions as anger \cite{Rhue_2018, Hitczenko_Cowan_Goldrick_Mittal_2022}. It is, therefore, unsurprising that language models trained on data reflecting hegemonic norms would over-perceive AAVE texts as angry and miss AAVE-written expressions of joy. Furthermore, we observe that model predictions directly perpetuate this stereotype when integrating Census data. In our Los Angeles Twitter dataset, the African American population density of a neighborhood correlates with higher predictions of anger and lower predictions of joy and love. Emotion AI's demographic correlation presents a significant risk that emotion AI systems may exacerbate unfavorable outcomes, including reduced access opportunities in the workplace \cite{Wingfield_2010} and poor communication with healthcare providers \cite{Park_Beach_Han_Moore_Korthuis_Saha_2020}. Between disproportionate false prediction rates on AAVE and race-based emotion correlates, emotion AI systems appear at risk of codifying race-based stereotypes. 

These results should inform more intentional applications of emotion detection systems as well as the consideration of whether the application of these systems is necessary in the first place. We encourage researchers and practitioners to critically engage with the sociocultural contexts of their study populations, particularly linguistic trends in emotional expression. As shown, the dissonance in understanding cultural linguistic features can significantly impact the interpretation and outcomes of emotion AI systems. Additionally, practitioners should seek to collect data annotations from the authoring communities of the text. Studies not focused on a clearly defined population should conduct linguistic diversity analyses to guide the selection of emotion labels and the metrics used to quantify affect presence and intensity. In conclusion, deployment of emotion AI systems on AAVE must be done with caution and a thorough understanding of both the model and the target data. 

\section{Limitations}
\paragraph{Annotations.}
Annotations on subjective tasks are inherently limited by those labeling them. We are interested in seeing how results shift with a different and larger set of annotators. Additionally, manual annotations are expensive. We would similarly like to see results on a different or a larger tweet sample.

\paragraph{Dialect Density.}
Dialects are constantly changing, with new expressions and constructs constantly being created, adopted, and dropped. Here, DDM approximates only those AAVE features grounded in sociolinguistics, resulting in a dialect density method inherently limited to those AAVE features covered by academic studies. Additionally, dialect density methods using named entity recognition or dependency parsing may be affected by noted biases in these methods \cite{blodgett-etal-2018-twitter}.

\paragraph{Flattening the Concept of Emotion.}
Emotions are a complex human phenomenon. Taking a discrete approach to modeling emotions inherently loses the nuance of emotions. By taking 20+ unique emotions and categorizing them into only 7 categories, we lose rich information. The original emotion web contains various intersections that become lost, maybe leading to both model and annotator disagreement purely on the categorical conceptualization.
% \chris{also that the original web that we start with has various intersections that we lose during this flattening which makes these categories clearly not orthogonal which can lead to both model an annotator disagreeance simply on the categories themselves }

\paragraph{Simplifying Race and Ethnicity.}
In comparing only two groups, African American dialect and communities with White or Standard dialect or communities, we miss out on Los Angeles' rich cultural diversity. A natural progression of our study is an expansion to Chicano American English, including its overlap with AAVE. 
Additionally, this work relies on definitions of race and ethnicity based on the United States Census. The conceptualization is heavily flawed and incomplete. However, Census-collected data provides arguably the most comprehensive demographic assessment that is publicly available. % We hope Census demographic classifications continue to expand to better incorporate the nuance of identity.

\section{Ethical Considerations}
\paragraph{Positionality Statement.}

The first author for this study is White and non-AAVE speaking. Two of the authors of this paper are Black and self-described AAVE-speaking. However, no faculty members are Black or AAVE-speaking. At the time of writing, all authors of this paper live in Los Angeles, California, with one of the AAVE-speaking authors growing up in Southern California. This work is informed by authors' past uses of emotion detection and frustration at its failings, as well as personal experiences being misclassified as aggressive.
%Conceptualizing African American speech as \textit{African American Vernacular English} yields questions towards whose language is being described and how the language variety is grounded in history \cite{King_2020}. 

\paragraph{Defining AAVE and GAE.}
AAVE is a single term that encompasses a wide variety of dialects based on region, gender, migration patterns and more. We attempt to narrow the variance by focusing on only the Los Angeles region. Inevitably, any conceptualization only represents select parts of the regional dialect.
Additionally, many assumptions are made when claiming a dialect as ``General American English". 
Following \cite{mugglestone2003talking}, we conceptualize GAE as a single dialect in order to ground group language within social norms and power.
%Here, we say Standard English to refer to the set of norms which share power across US regions.  

\paragraph{Relationship Between AAVE and Black Identity.}
We use demographic information as a signifier for increased likelihood of AAVE dialect. However, not all African American or Black individuals use AAVE, and not all who use AAVE are African American or Black. 

\paragraph{Environmental Impact of LLM Audits.}
Running large datasets through large language models requires a significant energy cost. We use model versions with the lowest parameters to try to offset our footprint. We hope the benefits of this paper justify the resource use.

%\paragraph{Privacy}
%We assume public tweet is consent to analyze
\begin{acks}
We would like to thank the creators of Latimer for their generosity.
We are grateful for Mary Kennedy, who provided invaluable linguistic expertise in our conceptualization of AAVE.
Thank you to Siyi Guo for helping with SpanEmo configurations.
Thank you to those at SoCalNLP who provided feedback on emotion model groundings and evaluation metrics: Katy Felkner, Jaspreet Ranjit, Leticia Pinto-Alva, and Arjun Subramonian.
\end{acks}

%%
%% The next two lines define the bibliography style to be used, and
%% the bibliography file.
\bibliographystyle{ACM-Reference-Format}
\bibliography{aaai25}

%%
%% If your work has an appendix, this is the place to put it.
\appendix

\begin{table*}[ht!]
\centering
\scalebox{0.95}{
\begin{tabular}{l|cc|cc|cc|cc|cc|cc|cc|cc}
\toprule
\multicolumn{1}{c}{ } & 
\multicolumn{2}{c}{\textbf{Joy}} & 
\multicolumn{2}{c}{\textbf{Love}} & 
\multicolumn{2}{c}{\textbf{Sadness}} & 
\multicolumn{2}{c}{\textbf{Anger}} &
\multicolumn{2}{c}{\textbf{Disgust}} &
\multicolumn{2}{c}{\textbf{Fear}} &
\multicolumn{2}{c}{\textbf{Surprise}} &
\multicolumn{2}{c}{\textbf{Overall}} \\
\cmidrule(lr){2-3}\cmidrule(lr){4-5}\cmidrule(lr){6-7}\cmidrule(lr){8-9}\cmidrule(lr){10-11}\cmidrule(lr){12-13}\cmidrule(lr){14-15}\cmidrule(lr){16-17}
\textbf{Model} & P & R & P & R & P & R & P & R & P & R & P & R & P & R & P & R \\
\toprule
\texttt{NRC Lexicon} & 0.28 & 0.20 & 0.15 & 0.24 & 0.21 & 0.19 & 0.27 & 0.15 & 0.07 & 0.12 & 0.02 & 0.47 & 0.06 & 0.12 & 0.15 & 0.21 \\
\texttt{RobertaGo} & 0.46 & 0.61 & 0.34 & 0.59 & 0.40 & 0.58 & 0.64 & 0.67 & 0.40 & 0.13 & 0.33 & 0.07 & 0.19 & 0.61 & 0.39 & 0.47 \\
\texttt{SpanEmo} & 0.42 & 0.89 & 0.34 & 0.69 & 0.28 & 0.68 & 0.49 & 0.91 & 0.12 & 0.95 & 0.02 & 0.67 & 0.19 & 0.41 & 0.27 & 0.74 \\
\midrule
\texttt{Deepseek-Qwen-zero} & 0.31 & 0.29 & 0.11 & 0.15 & 0.18 & 0.20 & 0.30 & 0.06 & 0.09 & 0.02 & 0.00 & 0.00 & 0.05 & 0.03 & 0.15 & 0.11 \\
\texttt{Deepseek-Qwen-few} & 0.30 & 0.21 & 0.12 & 0.22 & 0.24 & 0.27 & 0.25 & 0.03 & 0.11 & 0.02 & 0.01 & 0.20 & 0.07 & 0.05 & \textbf{0.16} & \textbf{0.14} \\
\texttt{Deepseek-Qwen-cot} & 0.28 & 0.19 & 0.08 & 0.10 & 0.24 & 0.29 & 0.23 & 0.03 & 0.07 & 0.02 & 0.02 & 0.20 & 0.00 & 0.00 & 0.13 & 0.12 \\
\midrule
\texttt{Llama-3.1-zero} & 0.30 & 0.36 & 0.13 & 0.15 & 0.22 & 0.15 & 0.27 & 0.24 & 0.09 & 0.18 & 0.02 & 0.08 & 0.02 & 0.03 & \textbf{0.15} & \textbf{0.17} \\
\texttt{Llama-3.1-few} & 0.29 & 0.35 & 0.15 & 0.19 & 0.13 & 0.08 & 0.30 & 0.27 & 0.06 & 0.12 & 0.02 & 0.07 & 0.05 & 0.07 & 0.14 & 0.16 \\
\texttt{Llama-3.1-cot} & 0.31 & 0.37 & 0.12 & 0.15 & 0.12 & 0.15 & 0.29 & 0.28 & 0.06 & 0.12 & 0.02 & 0.07 & 0.01 & 0.02 & 0.14 & 0.16 \\
\midrule
\texttt{GPT-4o-Mini-zero} & 0.55 & 0.89 & 0.59 & 0.51 & 0.67 & 0.31 & 0.65 & 0.84 & 0.46 & 0.69 & 0.23 & 0.44 & 0.20 & 0.35 & \textbf{0.48} & 0.58 \\
\texttt{GPT-4o-Mini-few} & 0.51 & 0.94 & 0.53 & 0.69 & 0.58 & 0.46 & 0.73 & 0.76 & 0.31 & 0.80 & 0.13 & 0.33 & 0.16 & 0.44 & 0.42 & \textbf{0.63} \\
\texttt{GPT-4o-Mini-cot} & 0.51 & 0.91 & 0.51 & 0.65 & 0.58 & 0.42 & 0.73 & 0.76 & 0.31 & 0.82 & 0.15 & 0.33 & 0.17 & 0.53 & 0.43 & \textbf{0.63} \\
\midrule
\texttt{Latimer-zero} & 0.53 & 0.91 & 0.58 & 0.55 & 0.62 & 0.30 & 0.68 & 0.81 & 0.38 & 0.53 & 0.14 & 0.33 & 0.18 & 0.33 & \textbf{0.45} & 0.54 \\
\texttt{Latimer-few} & 0.50 & 0.96 & 0.46 & 0.65 & 0.54 & 0.45 & 0.70 & 0.76 & 0.27 & 0.75 & 0.08 & 0.22 & 0.15 & 0.44 & 0.38 & 0.60 \\
\texttt{Latimer-cot} & 0.49 & 0.91 & 0.50 & 0.62 & 0.58 & 0.39 & 0.73 & 0.71 & 0.31 & 0.79 & 0.18 & 0.44 & 0.15 & 0.44 & 0.42 & \textbf{0.61} \\
\bottomrule
\end{tabular}
}
\caption{Precision (P) and Recall (R) across emotion categories and overall means. Values rounded to two decimals. For each generative model, highest precision and recall over all prompts is bolded. Tied values are both bolded.}
\label{tab:PR}
\end{table*}

\begin{table*}[h!]
\small
\begin{tabular}{c|ll|c}
\textbf{Feature} & \multicolumn{1}{c}{\textbf{Explanation}} & \multicolumn{1}{c|}{\textbf{Pattern}} & \textbf{Usage} \\ \hline

Abbreviations & 
\multicolumn{1}{l|}{\begin{tabular}[c]{@{}l@{}}Popular variations of \textit{I don't} and \textit{talking about} \\ from a study of AAVE regional patterns \cite{Jones_2015a}\end{tabular}} & 
\begin{tabular}[c]{@{}l@{}}\textcolor{blue}{\texttt{regex}}: \texttt{iont|iono|ioneem|sumn} \\ \texttt{|talmbout|talm bout}\end{tabular} & Introduce \\ \hline

Ain't & 
\multicolumn{1}{l|}{\begin{tabular}[c]{@{}l@{}}Expanded search string to reflect common variations \\ of the phrase \textit{aint} \cite{Jones_2015a}\end{tabular}} & 
\begin{tabular}[c]{@{}l@{}}\textcolor{blue}{\texttt{regex}}: \texttt{ain't|aint|yeen}\end{tabular} & Expand \\ \hline

Ass-Camo & 
\multicolumn{1}{l|}{\begin{tabular}[c]{@{}l@{}}Compound word with -ass suffix used as metonymic \\ pseudo-pronouns or discourse-level expressive markers \\ or intensifiers \cite{spears2013african}\end{tabular}} & 
\begin{tabular}[c]{@{}l@{}}\textcolor{blue}{\texttt{regex}}: \texttt{ass} \& \textcolor{green}{\texttt{DEP}}: [\texttt{POSS}, \texttt{DOBJ}] \\ | [\texttt{PUNCT}, \texttt{COMPOUND}] | [\texttt{AMOD}, \texttt{COMPOUND}]\end{tabular} & Introduce \\ \hline

Completive Done & 
\multicolumn{1}{l|}{\begin{tabular}[c]{@{}l@{}}Use of the word \textit{done} where the action is completed, \\ updated to POS \end{tabular}} & 
\begin{tabular}[c]{@{}l@{}}\textcolor{blue}{\texttt{regex}}: \texttt{done} \& \textcolor{red}{\texttt{POS}}: [\texttt{VERB}$_{i+1}$]\end{tabular} & Expand \\ \hline

Continuative Steady & 
\multicolumn{1}{l|}{\begin{tabular}[c]{@{}l@{}}Use of \textit{steady} to denote an action occurring \\ continuously and intensely\end{tabular}} & 
\begin{tabular}[c]{@{}l@{}}\textcolor{blue}{\texttt{regex}}: \texttt{steady} \& \textcolor{red}{\texttt{POS}}: [\texttt{NOUN}$_{i+1}$]\end{tabular} & Introduce \\ \hline

Copula Deletion & 
\multicolumn{1}{l|}{\begin{tabular}[c]{@{}l@{}}Absence of present-tense forms of copula \textit{be} \cite{stewart2014now, Rickford_2021}.\end{tabular}} & 
\begin{tabular}[c]{@{}l@{}}\textcolor{red}{\texttt{POS}}: \texttt{PRONOUN+ADJ}\end{tabular} & Introduce \\ \hline

Existential They & 
\multicolumn{1}{l|}{\begin{tabular}[c]{@{}l@{}}Terms such as \textit{am}, \textit{is}, and \textit{were} are omitted\end{tabular}} & 
\begin{tabular}[c]{@{}l@{}}N/A\end{tabular} & Remove \\ \hline

Habitual Be & 
\multicolumn{1}{l|}{\begin{tabular}[c]{@{}l@{}}\textit{Be} used to denote a habitual action, expanded to rely on \\ POS rather than perplexity differences \end{tabular}} & 
\begin{tabular}[c]{@{}l@{}}\textcolor{blue}{\texttt{regex}}: \texttt{be} \& \textcolor{red}{\texttt{POS}}: [\texttt{PRON}$_{i-1}$, \texttt{VERB}$_{i}$]\end{tabular} & Expand \\ \hline

N-Use & 
\multicolumn{1}{l|}{\begin{tabular}[c]{@{}l@{}}Usage of the n-word in positive or neutral manners \cite{spears2013african}\end{tabular}} & 
\begin{tabular}[c]{@{}l@{}}\textcolor{blue}{\texttt{regex}}: \texttt{(?i)nigga(s)?|n\textbackslash{}*gga(s)?}\end{tabular} & Introduce \\ \hline

Slang & 
\multicolumn{1}{l|}{\begin{tabular}[c]{@{}l@{}}Miscellaneous slang terms with at least 10K resulting \\ tweets from \cite{Jones_2015a}\end{tabular}} & 
\begin{tabular}[c]{@{}l@{}}\textcolor{blue}{\texttt{regex}}: \texttt{jawn|finna|doe|nawl|} \\ \texttt{nun|sholl|tryna|cuh}\end{tabular} & Introduce

\end{tabular}
\caption{All AAVE Dialect Density feature alterations from \citet{Johnson2022} base set. Entries show feature name, explanation of the dialect feature, the computational pattern elicited for feature presence, and, where applicable, how the feature differs from original methodology. In the Pattern column, \texttt{regex} denotes string match procedure, \texttt{DEP} denotes dependency parsing and \texttt{POS} means part of speech tagging labels using SpaCy \cite{spacy2025}. Where regex term checked for specific POS or NER value, subscript \textit{i} used to denote position of regex term in relation to pattern.}
\label{tab:ddm_feats}
\end{table*}

\end{document}